\documentclass[sigconf, 10pt]{acmart}

\usepackage{balance}
\usepackage{soul}
\usepackage{hyperref}
\usepackage{multirow}
\usepackage[utf8]{inputenc}
\usepackage[ruled, lined, longend, linesnumbered]{algorithm2e}
\usepackage{amsmath,multicol}
\usepackage{subcaption,textcomp,comment}
\usepackage{pifont}
\usepackage{tikz}

\usepackage{booktabs}
\usepackage{multirow}

\usepackage{paralist}
\usepackage{pifont}

\usepackage{ulem} 

\usepackage{array}

\usepackage{xcolor}
\definecolor{color0}{RGB}{0,0,0} 
\definecolor{color1}{RGB}{16,76,12} 
\definecolor{color2}{RGB}{105,2,20} 
\definecolor{color3}{RGB}{0,51,102} 
\definecolor{color4}{RGB}{149,41,107} 

\newcolumntype{L}[1]{>{\raggedright\let\newline\\\arraybackslash\hspace{0pt}}m{#1}}
\newcolumntype{C}[1]{>{\centering\let\newline\\\arraybackslash\hspace{0pt}}m{#1}}
\newcolumntype{R}[1]{>{\raggedleft\let\newline\\\arraybackslash\hspace{0pt}}m{#1}}

\newcommand{\mwx}[1]{\textbf{\color{red}{#1}}}

\newcommand{\cdq}[1]{{\color{blue}{#1}}}

\newcommand{\yjl}[1]{{\color{magenta}{#1}}}

\newcommand{\sys}{\texttt{M4}\xspace}
\newcommand{\benchmark}{\texttt{eAIBench}\xspace}

\def\circleA#1#2{\tikz[baseline=(char.base)]{
    \filldraw[fill=color0, draw=black] (0, 0.1) circle (0.24cm);
    \node[white, font=\bfseries, anchor=base] (char) at (0, 0.0) {\textbf{\scriptsize{#2}}};
    }}
    \def\circleAS#1#2{\tikz[baseline=(char.base)]{
    \filldraw[fill=color0, draw=black] (0, 0.1) circle (0.18cm);
    \node[white, font=\bfseries, anchor=base] (char) at (0, 0.04) {\textbf{\tiny{#2}}};
    }}
\def\circleB#1#2{\tikz[baseline=(char.base)]{
    \filldraw[fill=color1, draw=black] (0, 0.1) circle (0.24cm);
    \node[white, font=\bfseries, anchor=base] (char) at (0, 0.0) {\textbf{\scriptsize{#2}}};
    }}
    \def\circleBS#1#2{\tikz[baseline=(char.base)]{
    \filldraw[fill=color1, draw=black] (0, 0.1) circle (0.18cm);
    \node[white, font=\bfseries, anchor=base] (char) at (0, 0.04) {\textbf{\tiny{#2}}};
    }}
\def\circleC#1#2{\tikz[baseline=(char.base)]{
    \filldraw[fill=color2, draw=black] (0, 0.1) circle (0.24cm);
    \node[white, font=\bfseries, anchor=base] (char) at (0, 0.0) {\textbf{\scriptsize{#2}}};
    }}
    \def\circleCS#1#2{\tikz[baseline=(char.base)]{
    \filldraw[fill=color2, draw=black] (0, 0.1) circle (0.18cm);
    \node[white, font=\bfseries, anchor=base] (char) at (0, 0.04) {\textbf{\tiny{#2}}};
    }}
\def\circleD#1#2{\tikz[baseline=(char.base)]{
    \filldraw[fill=color3, draw=black] (0, 0.1) circle (0.24cm);
    \node[white, font=\bfseries, anchor=base] (char) at (0, 0.0) {\textbf{\scriptsize{#2}}};
    }}
    \def\circleDS#1#2{\tikz[baseline=(char.base)]{
    \filldraw[fill=color3, draw=black] (0, 0.1) circle (0.18cm);
    \node[white, font=\bfseries, anchor=base] (char) at (0, 0.04) {\textbf{\tiny{#2}}};
    }}
\def\circleE#1#2{\tikz[baseline=(char.base)]{
    \filldraw[fill=color4, draw=black] (0, 0.1) circle (0.24cm);
    \node[white, font=\bfseries, anchor=base] (char) at (0, 0.0) {\textbf{\scriptsize{#2}}};
    }}
    \def\circleES#1#2{\tikz[baseline=(char.base)]{
    \filldraw[fill=color4, draw=black] (0, 0.1) circle (0.18cm);
    \node[white, font=\bfseries, anchor=base] (char) at (0, 0.04) {\textbf{\tiny{#2}}};
    }}

\usepackage{wrapfig}
\usepackage{comment}
\usepackage{colortbl}

\usepackage{hyperref}

\usepackage{color,soul}
\usepackage{graphics}
\usepackage{hyperref}
\usepackage{lipsum}
\usepackage{tikz}
\usepackage{enumitem}	
\usepackage{listings} 

\usepackage{booktabs}	
\usepackage{lipsum}

\usepackage{capt-of}	

\usepackage{todonotes}  
\usepackage{subcaption} 

\definecolor{refkey}{rgb}{0,0,1}
\definecolor{labelkey}{rgb}{0,0,1}

\usepackage{cleveref}
\AtBeginDocument{\DeclareCaptionSubType{lstlisting}}
\crefname{sublstlisting}{listing}{listings}
\Crefname{sublstlisting}{Listing}{Listings}

\usepackage{mathrsfs}	
\usepackage{amsfonts}

\usepackage[export]{adjustbox} 

\usepackage{boldline}					

\usepackage{multirow}

\renewcommand{\paragraph}[1]{\vskip 3pt\noindent\textbf{#1 }}	 

  {\begin{list}{$\bullet$}%
     {\setlength{\parsep}{0pt}%
      \setlength{\topsep}{0pt}%
      \setlength{\itemsep}{2pt}}}%
  {\end{list}}
\newcommand\Noted[1]{} 

\definecolor{darkblue}{rgb}{0.0, 0.0, 0.55}
\definecolor{mygreen}{HTML}{ADFF2F}
\definecolor{mylightgray}{gray}{0.8}

  {\begin{itemize}
	[leftmargin=0cm,
		itemindent=.3cm,
		labelwidth=\itemindent,
		labelsep=0pt,
		parsep=0.0pt,
		topsep=0.5pt,
		itemsep=0.5pt,
		align=left]
  }%
  {\end{itemize}}    

  {\begin{enumerate}
	[leftmargin=.cm,itemindent=.5cm,labelwidth=\itemindent,
		labelsep=0pt,
		parsep=1pt,
		topsep=1pt,
		itemsep=3pt,
		align=left]
  }%
  {\end{enumerate}}    

\makeatletter
\def\@copyrightspace{\relax}
\makeatother

\begin{document}

    \title[Mobile Foundation Model as Firmware]{Mobile Foundation Model as Firmware\\
    \small The Way Towards a Unified Mobile AI Landscape}
    \author{Jinliang Yuan$\dag$, Chen Yang$\dag$, Dongqi Cai$\dag$, \\ Shihe Wang, Xin Yuan, Zeling Zhang, Xiang Li, Dingge Zhang, Hanzi Mei, Xianqing Jia, Shangguang Wang, Mengwei Xu*}
	\thanks{$\dag$Authors contributed equally to this research. *Corresponding author.}
    \affiliation{
	  \institution{Beijing University of Posts and Telecommunications}
	  \city{Beijing}
	  \country{China}
	}
    \renewcommand{\shortauthors}{Jinliang Yuan et al.}
    
    \begin{abstract}  
In the current AI era, mobile devices such as smartphones are tasked with executing a myriad of deep neural networks (DNNs) locally.
It presents a complex landscape, as these models are highly fragmented in terms of architecture, operators, and implementations. Such fragmentation poses significant challenges to the co-optimization of hardware, systems, and algorithms for efficient and scalable mobile AI.
	
Inspired by the recent groundbreaking progress in large foundation models, this work introduces a novel paradigm for mobile AI, where mobile OS and hardware jointly manage a foundation model that is capable of serving a wide array of mobile AI tasks. 
This foundation model functions akin to firmware, unmodifiable by apps or the OS, exposed as a system service to Apps. 
They can invoke this foundation model through a small, offline fine-tuned ``adapter'' for various downstream tasks.
We propose a tangible design of this vision called \sys, and prototype it from publicly available pre-trained models.
To assess its capability, we also build a comprehensive benchmark consisting of 38 mobile AI tasks and 50 datasets, spanning 5 multimodal inputs.
Extensive experiments demonstrate \sys's remarkable results: it achieves comparable accuracy in 85\% of tasks, offers enhanced scalability regarding storage and memory, and has much simpler operations. In broader terms, this work paves a new way towards efficient and scalable mobile AI in the post-LLM era.
	


\end{abstract}

    \acmYear{2024}\copyrightyear{2024}
    \acmConference[ACM MobiCom '24]{International Conference On Mobile Computing And Networking}{September 30--October 4, 2024}{Washington D.C., DC, USA}
    \acmBooktitle{International Conference On Mobile Computing And Networking (ACM MobiCom '24), September 30--October 4, 2024, Washington D.C., DC, USA}
    \acmDOI{10.1145/3636534.3649361}
    \acmISBN{979-8-4007-0489-5/24/09}

    \maketitle
    
\section{Introduction}\label{sec:intro}

Machine learning is revolutionizing mobile applications by facilitating a more automated, intelligent, and efficient interaction between users and devices. These advancements enable humans to enjoy the convenience provided by deep models at all times and locations, from voice assistants \cite{granqvist2020improving, websiri}, image editers \cite{sun2021semantic, wang2023images, xu2023imagereward}, to augmented reality \cite{prakash2019gleam, zhao2021xihe}. As reported in~\cite{almeida2021smart,xu2019first}, the number of deep models incorporated within individual devices is growing rapidly, making mobile devices a primary vehicle for AI.

Executing deep models on devices offers benefits in data privacy and service availability but also demands significant resources such as memory, energy, and time. For efficient and scalable on-device execution of these models, a comprehensive co-design approach that integrates hardware, system, and algorithms is needed.
However, this task is challenged by the \textit{fragmented ecosystem} of mobile deep models:
they significantly differ in architecture, operators, and implementations \cite{lane2016deepx, georgiev2016leo, huynh2017deepmon, yang2013beyond, sun2020mobilebert, chen2022mobile}.
This fragmentation, which often results in ad-hoc optimization efforts \cite{alwani2016fused, pati2022demystifying, shen2017maximizing}, seems unavoidable.
It originates from the complex nature of mobile AI tasks (CV/NLP/TTS/HAR/..), multimodal data from various sensors (camera, screen, microphone, etc.), and diverse application demands (high accuracy, low latency, etc.).

Such fragmentation fundamentally undermines the efficiency of constructing an efficient and scalable mobile AI stack, notably in the three following aspects:

\begin{figure}[t]
	\centering
	\includegraphics[width=0.48\textwidth]{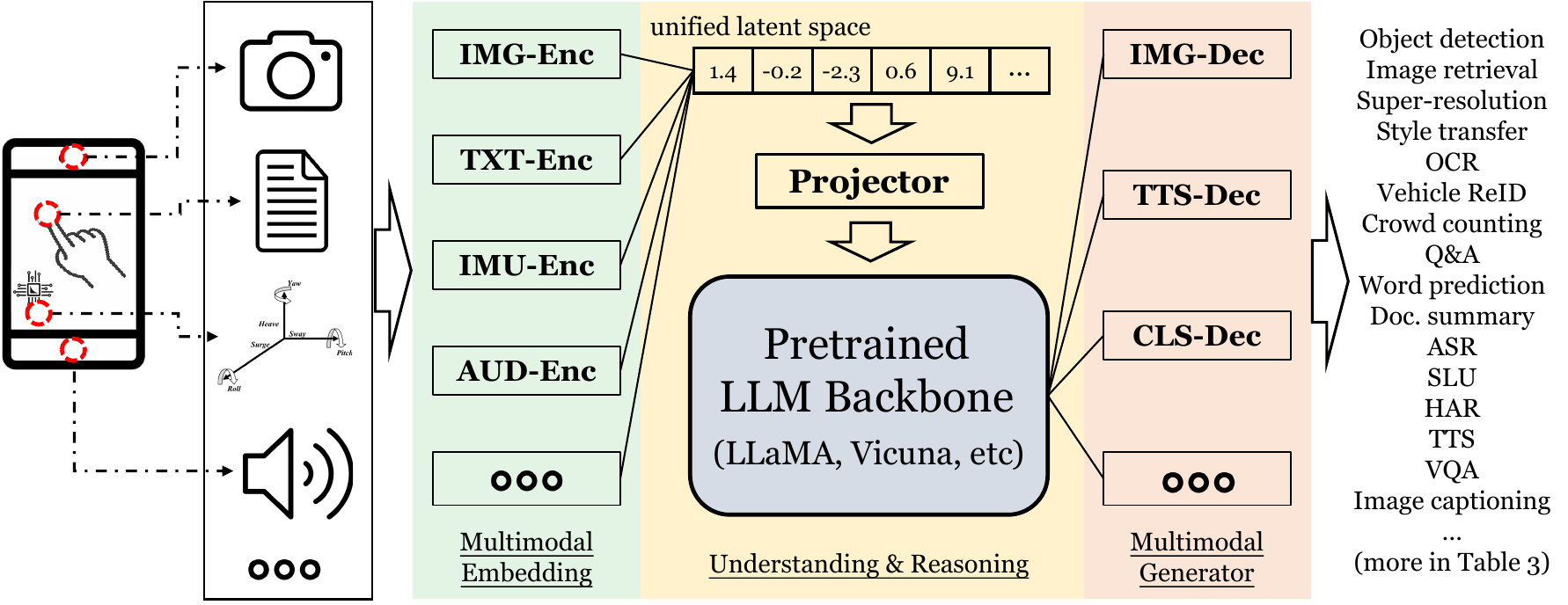}
	\caption{An overview of \sys.
	}
	\label{fig:MobileFM-archi}
	\vspace{-10pt}
\end{figure}

\begin{itemize}[leftmargin=*,topsep=1pt]
	\item \textit{Hardware aspect:} it complicates the design of ASIC-based accelerators (NPUs), by forcing difficult trade-offs between generality and performance.
	$\S$\ref{subsec:bkgnd-heterNN} shows that mobile NPUs can achieve up to a 22$\times$ speedup over multi-core CPUs on the Qualcomm Snapdragon 8+ Gen 1 platform. However, this advantage only extends to a small fraction (around 8\%) of deep models due to the lack of operator support.
	
	\item \textit{OS aspect:} It hampers the system-wise sharing of weights and computations across different applications. 
	Mobile apps often perform similar meta-ML tasks (e.g., object detection for augmented reality, image enhancement, OCR apps), and there exist temporal, spatial, and semantic correlations among the input data \cite{guo2018potluck}.
	However, exploiting such similarities to reduce memory or computing via cache-reuse is currently impractical, due to the model fragmentation and OSes' lack of visibility into those models managed at the application level.
	
	\item \textit{Software aspect:} It makes library-level optimizations ad-hoc.
	As noted in \cite{zhang2022comprehensive}, there is a wide array of frameworks available to developers, but their performance can vary significantly across different models and devices. No single solution excels universally, often leaving developers struggling to differentiate between them. 
\end{itemize}

\noindent \textbf{Mobile foundation model as firmware.}
In order to fundamentally tackle the aforementioned issues, we propose a novel paradigm for mobile AI in which the OS and hardware co-manage a foundation model that is capable of addressing most, if not all, mobile AI tasks. This model, akin to firmware, 
is exposed as a system service to applications similar to the unified ML interface NNAPI~\cite{NNAPI2017} in Android.
It remains unaltered by apps or the OS. To utilize it, each application embeds it a lightweight "adapter" that is fine-tuned offline for downstream tasks. This approach could greatly simplify NPU design 
and allow the OS to take control of AI computing across applications, thereby facilitating the sharing of weights, computations, and task scheduling.
This vision becomes feasible thanks to recent advancements in the ML community, specifically:
(1) The establishment of pre-trained foundation models \cite{touvron2023llama, GPT4, scao2022bloom} that capture extensive knowledge from vast Internet data;
(2) The development of algorithms to accurately align multimodal data input \cite{girdhar2023imagebind, tang2023Codi};
(3) The demonstration of parameter-efficient fine-tuning (PEFT) methods like LoRA \cite{hu2021lora, peft2023} that efficiently adapt pre-trained models to diverse downstream tasks.

While the vision is intriguing, there are two key missing pieces to turn it into reality.
(1) How to build such a one-size-fits-all foundation model to ubiquitously handle highly diversified, multimodal mobile AI tasks?
While research on the multimodal foundation model has achieved impressive progress in recent years, they are still not adequate in our case:
most of them~\cite{li2021align,najdenkoska2023meta,radford2021learning} handle only a small fixed number of input/output modalities (e.g., text-image) and cannot be flexibly adapted to more;
a concurrent effort CoDi~\cite{tang2023Codi} with this work enables any-to-any generation across three modalities (image-text-audio), but requires more than 34GBs of on-device storage/memory.
(2) How to properly evaluate the performance of the proposed foundation model?
To our best knowledge, there has been no comprehensive benchmark or a set of standard metrics for mobile AI tasks.

\noindent \textbf{\sys: a composable mobile foundation model ($\S$\ref{sec:design-model}).}
We introduce \sys, the first architectural design and implementation of a multimodal mobile foundation model, as illustrated in Figure~\ref{fig:MobileFM-archi}.
Unlike prior approaches like CoDi that directly use (Nx) heavy encoders to align multimodal input data and (Mx) heavy decoders to generate specific data format,
\sys adds a backbone module in between (a ``narrow waist'') that comprehends and reasons for each downstream task.
Through such ``N-1-M'' design, \sys is able to achieve better accuracy-to-parameter efficiency as compared to traditional ``N-M'' architecture.
Moreover, \sys could be partially activated by various tasks based on their characteristics (input/output modality, the need for complex comprehension, etc.).
We have fully prototyped \sys with only pre-trained models publicly available from HuggingFace~\cite{wolf2019huggingface}, which guarantees the reproducibility of \sys and also demonstrates its compatibility with the existing LLM ecosystem.
Overall, \sys contains 9.2B parameters and demands 7.5GBs of peak memory footprint.
Such a size is only affordable on high-end mobile devices nowadays, but we deem it to be soon feasible for more commons whose memory/storage capacity is significantly increasing yearly.

\noindent \textbf{\benchmark: a comprehensive edge-oriented AI benchmark ($\S$\ref{subsec:eval-setup}).}
To assess \sys and future endeavors, we have constructed the first comprehensive benchmark for diverse mobile AI tasks, named \benchmark.
Through an extensive examination of real-world mobile AI and publications in mobile venues, \sys presently includes 38 important mobile AI tasks and 50 classic datasets.
The tasks include five different input/output data modalities (vision, text, audio, IMU, and mix).
Each task is also linked with a task-specific model, representative of the DNN in the pre-LLM era (e.g., ResNet-152 for image classification \cite{he2016deep} and LSTM for input token prediction \cite{huang2015bidirectional}). 
We also standardize a set of key metrics to quantify the capability of a foundation model.

\noindent \textbf{Key results ($\S$\ref{sec:exps}).}
We then conduct extensive experiments to evaluate \sys using \benchmark on three kinds of hardware platforms: NVIDIA A100 GPU, NVIDIA Jetson ORIN NX and Pixel 7 Pro smartphone.
We summarize our major results.

$\bullet$ \textit{\textbf{Ubiquity--} \sys effectively supports most tasks and datasets in \benchmark.}
Compared with the models tailored for each task, \sys shows comparable accuracy on 85\% of the 50 datasets and a significant improvement on 4 of them (including image captioning and text-to-image retrieval).
In only six instances does \sys experience nontrivial accuracy degradation, marked by a greater than 10\% gap. The system also demonstrates promising zero-shot and few-shot capabilities, achieving usable accuracy on certain tasks without any fine-tuning. Moreover, quantization minimally affects the performance of \sys: when reduced to 8 bits on two tested tasks, accuracy degradation ranges only between 0.2\% and 0.8\%.
To be noted, the backbone LLM used in the current prototype of \sys, i.e., LLaMA (Feb. 2023), has been defeated by many other open-source LLMs since its release, such as LLaMA-2 (July. 2023) and Mistral-7B~\cite{jiang2023mistral} (Oct. 2023).
We expect the performance of \sys to improve substantially as well by using such more powerful backbone LLMs.
This is also confirmed by our preliminary experiments by replacing LLaMA with LLaMA-2 on two tasks, as will be discussed in $\S$\ref{subsec:accuracy}.

$\bullet$ \textit{\textbf{Scalability --} Despite \sys foundation model's heavier footprint, its adaptation to downstream mobile tasks is lightweight and therefore more scalable.}
The current implementation of \sys encompasses  $\sim$10 billion parameters, in contrast to the mere 1 million to 500 million parameters found in task-specific models. Nevertheless, the "adapters" of \sys require only 1,000 to 10 million parameters, which enhances scalability across various mobile AI tasks, given that the foundation model is shared. For example, on a device with 12GB of memory, \sys (4-bit quantized) with all 50 adapters can be hosted in memory, eliminating cold-start latency, whereas only 20 of 50 task-specific models would fit the same memory constraints.

$\bullet$ \textit{\textbf{Velocity --} 
    \sys is much slower than task-specific models, yet the gap might be mitigated through a highly-optimized NPU.}
On a high-end autonomous board Jetson Orin NX (16GB memory), \sys runs 18$\times$ slower on average.
We also test the performance of \sys on smartphone CPUs
\footnote{Currently, \sys cannot run on COTS smartphones GPU/NPU due to the lack of operator support.}
, which shows that the prediction delay could be too slow, i.e., 2.1 secs to classify an image or 240 msecs to generate a token in QA.
However, such a performance degradation might be addressed by running \sys on a highly optimized NPU, since existing NPUs already offer up to a 22-fold speedup over CPUs, as mentioned in $\S$\ref{subsec:bkgnd-heterNN}.

$\bullet$ \textit{\textbf{Simplicity --} \sys requires fewer operators for execution, greatly simplifying hardware design.}
In the ONNX format, \sys utilizes a mere 39 different mathematical operators, in contrast to the cumulative 156 operators required by 50 task-specific models. More impressively, \sys can expand its capabilities using the same number of operators. The traditional approach, on the other hand, continuously introduces new operators \cite{yu2022orca, zhou2023mpress}, thereby complicating NPU design.

In addition to conventional mobile AI tasks, \sys also enables more complex and innovative mobile applications, e.g., a sophisticated assistant capable of processing multimodal input data, understanding user intentions, and responding with precision as demonstrated in $\S$\ref{subsec:usecase}.


\noindent  \textbf{Contributions} Major contributions are summarized below.
\begin{itemize}[leftmargin=*,topsep=2pt]
\item We delineate a vision for a mobile foundation model, harnessing cutting-edge machine learning techniques to consolidate the mobile AI ecosystem and foster integrated hardware-system co-design.

\item We design and prototype the first mobile foundation model with public, pre-trained LLMs.

\item We have constructed the first comprehensive edge-oriented AI benchmark, through which our prototype demonstrates significant potential in catering to widespread mobile AI tasks, while exhibiting strong scalability, flexibility, and velocity in its performance.
\end{itemize}

\noindent \textbf{Open-source}
\sys and \benchmark are publicly available at \url{https://github.com/UbiquitousLearning/MobileFM}.

    \section{Background and Motivation} \label{sec:bkgnd}

\subsection{Mobile AI Characteristics}
\label{subsec:bkgnd-pervasive}

\paragraph{Mobile AI is pervasive.}
An important trend of AI deployment is the migration of deep learning inference tasks from data centers to smartphones, aiming to minimize user-perceived latency and better preserve data privacy \cite{chen2022mobrecon, cui2022dvabatch}. 
For instance, it is reported that Android apps embedded with on-device DNNs on the Google Play market have experienced a remarkable 60\% growth from February 2020 to April 2021~\cite{almeida2021smart};
Such DL-enhanced apps have been downloaded by users billions of times.
Unsurprisingly, mobile devices like smartphones and laptops have become a major carriers of intelligence, where DNN inference happens frequently anywhere anytime even without users being aware of it.

\paragraph{Mobile DNNs are fragmented.}
Unlike cloud AI where each computing unit (e.g., an NVIDIA GPU) only serves one model for user requests \cite{shen2019nexus, geoffrey2021habitat, muthukrishnan2021efficient}, a mobile device needs to handle highly diversified mobile AI tasks by itself.
Such diversification is inevitable since mobile AI tasks could leverage multimodal sensor data from devices, including imagery data from cameras, audio data from microphones, IMU data from motion sensors, and textual/code data from users typing.
Each modality itself has a wide spectrum of applications, e.g., Google Translate for NLP and Apple Siri for Audio.
Meanwhile, there are a wide range of cross-modal applications in mobile scenarios: visual question answering \cite{zhang2023vqacl}, image captioning \cite{cornia2019show}, and multimedia content retrieval \cite{liu2021hit}.
Research suggests that the quantity of multimodal applications on mobile devices has almost doubled in the last two years due to rapid advancements in multimodal technologies \cite{yin2023survey}. A recent empirical in-the-wild study \cite{almeida2021smart} reported significant heterogeneity in the architectures and internal operators of DNNs handling various modal data.

\subsection{A Dilemma of Mobile NPU}
\label{subsec:bkgnd-heterNN}

Fragmented DNNs significantly strain mobile AI stack hardware, system, and library design, as discussed in Section \ref{sec:intro}. Here, we emphasize the challenges faced by mobile NPUs specifically. Pilot measurements are conducted to assess mobile NPUs' performance gains compared to mobile CPU/GPU when executing typical mobile DNNs.

\begin{figure}[t]
	\centering
	\includegraphics[width=0.48\textwidth]{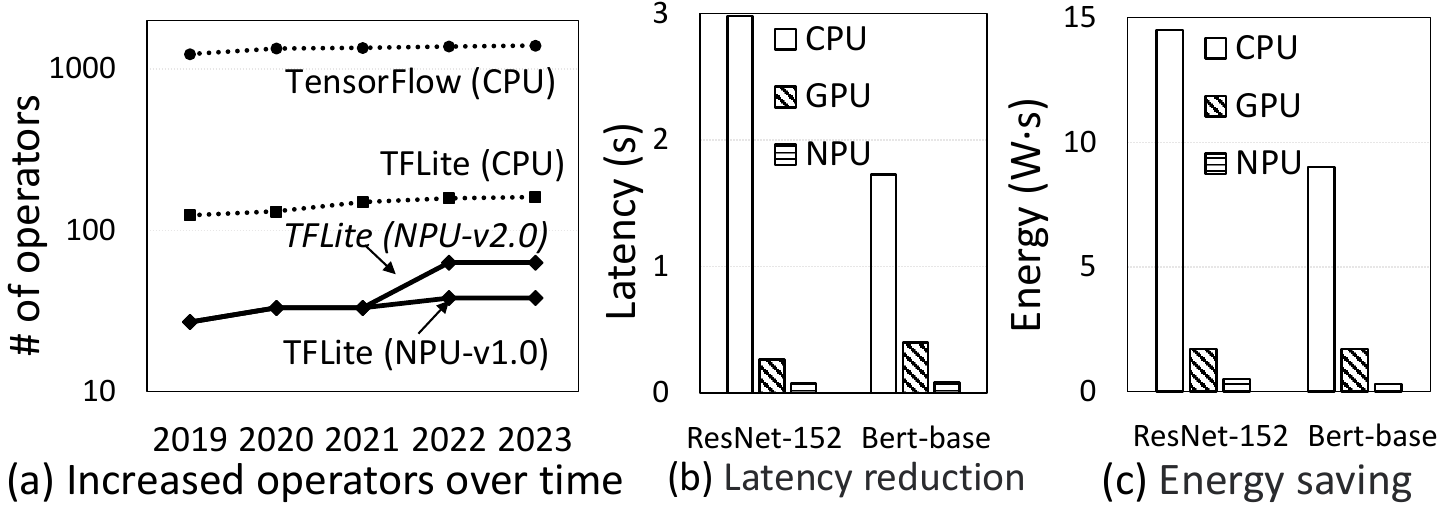}
	\caption{Empirical study on mobile NN processors:
		(1) A longitudinal analysis of operator support on CPU/NPU; 
		(2) The performance gap between NPU and CPU/GPU on Google Pixel 7 Pro.
	}
	\label{fig:npu-operators}
\end{figure}

\paragraph{The DNN operator support of mobile NPU is significantly lagged behind general-purpose processors.
}
We conducted an investigation on the number of supported NN operators by TensorFlow and TFLite.
As illustrated in Figure~\ref{fig:npu-operators} (a), we have two key observations:
(1) The number of NN operator types is still increasing noticeably lately, e.g., from 1240 to 1399 as supported by TensorFlow from 2019 to 2023.
Such evolvement of DNN architecture poses significant challenges in designing ubiquitous and efficient mobile NPU design.
(2) Mobile chips, especially its NPU, support only a small portion of existing NN operators.
TFLite supports less than 160 operators on mobile CPUs, which is nearly 90\% fewer than TensorFlow.
Furthermore, the number of supported operators by mobile NPU (EdgeTPU on Pixel 7 Pro) is even fewer, i.e., 33 in 2022 and 63 in 2023. 
Consequently, mobile NPUs might benefit only a small number of DNNs.

\paragraph{For the lucky DNNs fully supported, mobile NPU is able to deliver significant inference speedup and energy reduction compared to mobile CPU/GPU.}
As an ASIC-based customized processor, mobile NPU is expected to offer faster and more energy-efficient DNN inference.
To understand the performance of contemporary mobile NPU, we measure the inference latency and energy consumption of EdgeTPU on Google Pixel 7 Pro.
The results are illustrated in Figure~\ref{fig:npu-operators} (b) and (c).
ResNet-152, the NPU achieves an inference latency of only 76ms, which is 39$\times$ and 11$\times$ faster than the accompanying CPU (4-cores used) and GPU, respectively.
Similarly, on BERT-base, the NPU consumes only 0.3J of energy per image, while the CPU and GPU consume significantly higher amounts of energy, i.e.,  1.7J (5.78$\times$) and 9.0J (30$\times$), respectively.

\begin{figure}[t]
	\centering
	\includegraphics[width=0.45\textwidth]{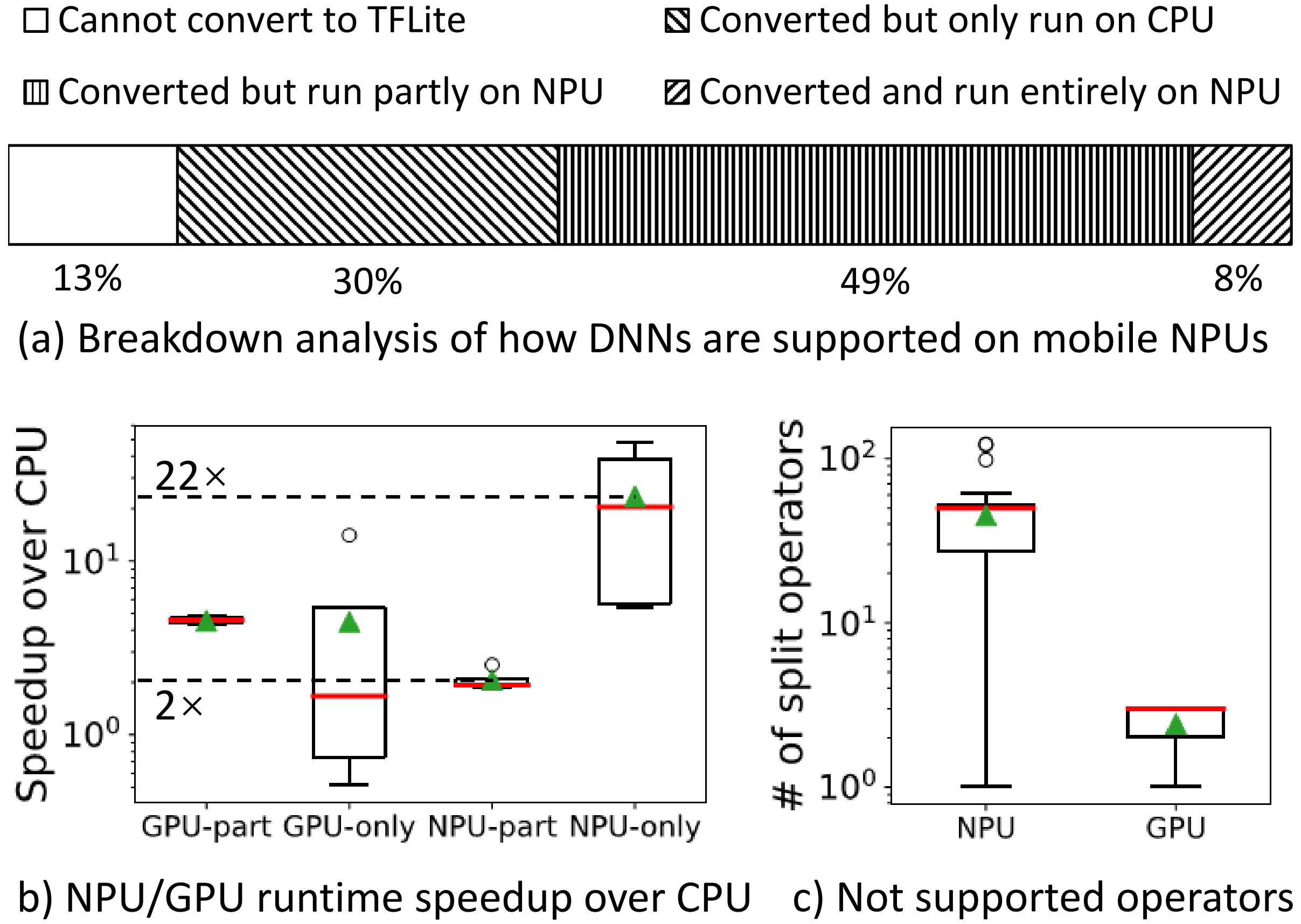}
	\caption{An empirical study of 110 in-the-wild DNNs crawled from public sources on Google Pixel 7 Pro.
	}
	\label{fig:heterogeneous-operators}
\end{figure}

\paragraph{However, such significant benefits only apply to a very small portion of in-the-wild popular DNNs. 
}
To further understand the ubiquity of mobile NPU, we download more representative DNNs and test their performance.
In total, we have found 110 TensorFlow-format DNNs from Model Zoo \cite{Modelzoo} and HuggingFace \cite{Huggingface}, prioritized based on their stars and download times.
We then try to convert them to TFLite format using the official tool developed by Google and measure their performance on Google Pixel 7 Pro.
As shown in Figure~\ref{fig:heterogeneous-operators} (a), unfortunately, only 8\% of those models can entirely run on mobile NPU, while for the rest:
13\% fail to be converted to TFLite format;
30\% fail to run on mobile NPU;
and 49\% require CPU-NPU co-running due to the lack of NN operator support by NPU.
Figure~\ref{fig:heterogeneous-operators} (b) further illustrates the performance of those DNNs on devices.
It reveals that,
only the 8\% fully supported DNNs gain significant improvement over CPUs (i.e., $>20\times$ median speedup), while the rest (either running partly on GPU/NPU or entirely on GPU) obtain much less profound speedup.
In fact, for the DNNs that require CPU-NPU co-running, the inference speed is even not as good as running on mobile GPU.
Figure~\ref{fig:heterogeneous-operators} (c) further digs into the reason for such phenomenon:
the NPU-incompatible DNNs need to be split into many sub-models to be scheduled between CPU and NPU (e.g., median number of 50); therefore the data movement and format exchange could severely delay the inference.

\subsection{Emergence of Foundation Models}
\label{subsec:bkgnd-LLM}


\paragraph{Foundation models are renovating AI ecosystem; the model design is converging.}
In recent years, significant advancements have been made in large-scale neural networks for language, image, and audio understanding and generation. GPT-3 \cite{brown2020gpt3} exemplifies this progress with impressive performance across various tasks, revolutionizing human-computer interaction and intelligent assistance. In the visual domain, Meta's SAM \cite{kirillov2023segment} demonstrates exceptional zero-shot proficiency. Additionally, models like Kosmos-1 \cite{huang2023language} and PaLM-E \cite{chowdhery2022palm} handle inputs from multiple modalities, enabling diverse task capabilities. These models share the transformer architecture \cite{vaswani2017attention}, differing mainly in layer configurations or input modality processing. This convergence trend in AI model design is expected to continue in the future.

\paragraph{However, there has been no effort in building one model to fit highly diversified mobile AI tasks.}
None of the aforementioned foundation models is capable of (not even close to) solving all mobile AI tasks.
A single modality model (such as GPT for NLP) cannot comprehend or generate data in other modalities.
Existing multimodal models (such as CLIP for CV-NLP) can only deal with very limited multimodal AI tasks.
One might seek to include a foundation model for each $<input:M1, output:M2>$ pair to solve the above issue, but:
(1) It is not parameter-efficient as the comprehension and conversion between different modality data share inherent common sense~\cite{li2021align,radford2021learning};
(2) It cannot support AI tasks that take multimodal input or output, such as visual question answering \cite{zhang2023vqacl}.
There have been ad-hoc approaches to deal with those issues \cite{su2023pandagpt}, yet we are not aware of any systematic strategy to build a one-size-fits-all foundation model for diversified mobile AI tasks.

\section{\sys Design and Prototyping}\label{sec:design-model}



\subsection{Overview}
\label{sec:design-overview}

\paragraph{Design principles}
\sys is a one-size-fits-all foundation model for diversified mobile AI tasks.
It is designed with following principles:
(1) \textit{unified}: instead of building independent foundation models for different possible modalities, \sys provides a unified architecture that maximizes the capability sharing across different modalities, thus being more resource-efficient and extensible;
(2) \textit{elastic}: \sys can be easily scaled out to more modalities (either for input or output), e.g., for new types of sensor/app data;
(3) \textit{multimodal}: \sys can take multimodal input or generate multimodal output as needed, e.g., for advanced mobile applications like visual question answering or audio caption.

\paragraph{Model architecture}
Figure~\ref{fig:MobileFM-archi} illustrates the overall architecture of \sys, which consists of three major components:

\noindent $\bullet$ \textit{Multimodal Embedding} 
is to align the contents of different modalities by converting multimodal input data into a unified representation (i.e., a vector).
It is typically implemented as a set of transformer encoders~\cite{girdhar2023imagebind} for each modality, except that audio has two independent encoders to differentiate the context information (e.g., background noise, speaker emotions) and spoken language (e.g., automatic speech recognition).

\noindent $\bullet$ \textit{Foundation Backbone (i.e., Pre-trained LLM Backbone)} is to comprehend and reason about the input data.
It encapsulates abundant knowledge to understand complex embedded data, performs task-specific inference, and generates easily intelligible output for generator. 
It uses a decoder-based architecture trained on huge amount of textual dataset since language has been acknowledged as the most representative type of data~\cite{GPT4,devlin2018bert,touvron2023llama}.
The backbone is the most heavy part of \sys.

\noindent $\bullet$ \textit{Multimodal Generator}  is to adapt the output from the foundation backbone to task-specific data format.
For classification tasks, it is simply a MLP with softmax layer;
for image tasks, it is a stable diffusion model~\cite{rombach2022high}; etc.

\paragraph{Trainable parameters}
\sys contains three trainable parts to be fine-tuned for downstream mobile AI tasks:
two PEFT modules inserted to the multimodal embedding and foundation backbone, respectively;
and one MLP projection layer that adapts the output of multimodal embedding to the required representation of the foundation backbone.
In later experiments, we use LoRA as the default PEFT method, but also report results for other PEFT methods.
As will be demonstrated in $\S$\ref{sec:exps}, the trainable parameter size is trivial compared to the pre-trained part of \sys and is also much smaller than traditional state-of-the-art DNNs.

\subsection{Prototyping with Off-the-Shelf LLMs}
\label{subsec:impl}

We have fully prototyped \sys with only pre-trained, off-the-shelf models publicly available from HuggingFace~\cite{wolf2019huggingface}.
It guarantees the reproducibility of \sys and also demonstrates its compatibility with the existing LLM ecosystem.

\noindent $\bullet$ \textit{Multimodal Embedding.} 
Multimodal embedding is composed of five parallel modules with transoformer encoder-only architecture: \textit{Image (IMG\_{enc})}, \textit{Text (TXT\_{enc})}, \textit{Inertial Measurement Unit (IMU\_{enc})}, \textit{Audio-Background (AUD-B\_{enc})}, and \textit{Audio-Intent (AUD-I\_{enc})}.
The \textit{IMG\_{enc}} employs the Vision Transformer (ViT) architecture and is utilized to encode visual information derived from input images into a sequential matrix of embeddings. 
The \textit{TXT\_{enc}} for input text is based on the CLIP architecture with a 12-layer encoder~\cite{radford2021learning}. 
The \textit{IMU\_{enc}} for IMU data is a lightweight 6-layer encoder transformer model~\cite{radford2021learning}.
The \textit{AUD-B\_{enc}} encoder is also derived from ViT and is used for encoding audio backgrounds~\cite{gong2021ast}. 
The pre-trained weights of the above four encoders are from ImageBind~\cite{girdhar2023imagebind} multimodal model.
The \textit{AUD-I\_{enc}} encoder is based on a sequence-to-sequence Transformer model for encoding audio intents, with pre-trained weights from Whisper.tiny.en~\cite{radford2023robust}.

\noindent $\bullet$ \textit{Foundation Backbone.} We use LLaMA-7B (INT8 format)~\cite{touvron2023llama}, pre-trained on one trillion tokens by Meta, as \sys's backbone.
Released in Feb. 2023, LLaMA is a research project aimed at creating a more versatile and efficient language model.
It emphasizes training on a broad array of multilingual and multitask supervised data to enhance performance across various natural language processing tasks.

\noindent $\bullet$ \textit{Multimodal Generator.}
Multimodal generator is composed of three parallel decoders: \textit{Text-to-Speech (TTS\_{dec})}, \textit{Image (IMG\_{dec})} and \textit{Generation (GEN\_{dec})}.
The \textit{TTS\_{dec}} decoder is a integral element within the FAIRSEQ~\cite{ott2019fairseq}, the open-source sequence modeling toolkit released by Meta, tasked with converting the input text into corresponding speech signals.
The \textit{IMG\_{dec}}  decoder is a key component of the Diffusion Model, which generates image output from text input.
The \textit{GEN\_{dec}} decoder serves as distinctive entities employed to lead the backbone language model to perform generation tasks, the parameters of which are initialized with the last layer of pre-trained LLaMA.
Classification tasks could be reformulated with a generation prompt according to prompt learning literature~\cite{liu2021pre} and re-use the generation decoder \textit{GEN\_{dec}}.
Or it could re-initialize a new MLP decoder according to traditional classification literature.


\begin{table}[t]
    \centering
    \resizebox{0.96\columnwidth}{!}{%
    \begin{tabular}{|c|c|c|c|c|c|}
    \hline
                                      & \textbf{Types}   & \textbf{Params ($10^9$)} & \textbf{Format} & \textbf{Architecture} & \textbf{GFLOPs} \\ \hline
    \multirow{5}{*}{\textbf{Embedding}} & IMG\_enc~\cite{girdhar2023imagebind}              & 0.6328       &FP16               & Encoder-only                      & 167.5963                     \\ \cline{2-6} 
                                      & TXT\_enc~\cite{girdhar2023imagebind}              & 0.354                 &FP16       & Encoder-only                      & 23.4189                      \\ \cline{2-6} 
                                      & AUD-B\_enc~\cite{girdhar2023imagebind}            & 0.0862               &FP16        & Encoder-only                      & 61.4679                      \\ \cline{2-6} 
                                      & AUD-I\_enc~\cite{radford2023robust}           & 0.037                &FP16        & Encoder-Decoder                         & 26                           \\ \cline{2-6} 
                                      & IMU\_enc~\cite{girdhar2023imagebind}              & 0.0196             &FP16          & Encoder-only                       & 5.1417                       \\ \hline
    \textbf{Backbone}                 & LLaMA~\cite{touvron2023llama}           & 6.28             &INT8               & Decoder-only                         & 312                          \\ \hline
    \multirow{3}{*}{\textbf{Generator}} & TTS\_dec              & < 0.01               &FP32     & Encoder-Decoder                    & 8.58                         \\ \cline{2-6} 
                                      & IMG\_dec~\cite{rombach2022high} & 1.0663               &FP16       & Encoder-Decoder                        & 267                       \\ \cline{2-6}
                                      & GEN\_dec~\cite{touvron2023llama} & < 0.01               &FP16       & MLP                        & 125.0                       \\ \cline{2-6}
                                        \hline
    \end{tabular}%
    }
    \caption{\sys sub-model parameters. 
    }
    \label{tab:design-fm-size}
    \vspace{-20pt}
    \end{table}

\paragraph{System complexity}
Table~\ref{tab:design-fm-size} presents the model complexity of \sys's different modules.
The model comprises five types of embeddings (encoders), with parameter sizes ranging from 0.03B to 0.63B, and complexities ranging from 26GFLOPs to 167GFLOPs. 
The backbone of \sys is LLaMA, with a parameter count of 6.3B, and a complexity of 312GFLOPs. 
The LLaMA backbone is the largest component (86.1\%) in terms of parameter size within the entire model.
The generators (decoders) contribute trivially to the overall model size.

\subsection{Multi-path Task Execution}
\label{sec:design-config}

\begin{figure}[t]
	\centering
	\includegraphics[width=0.48\textwidth]{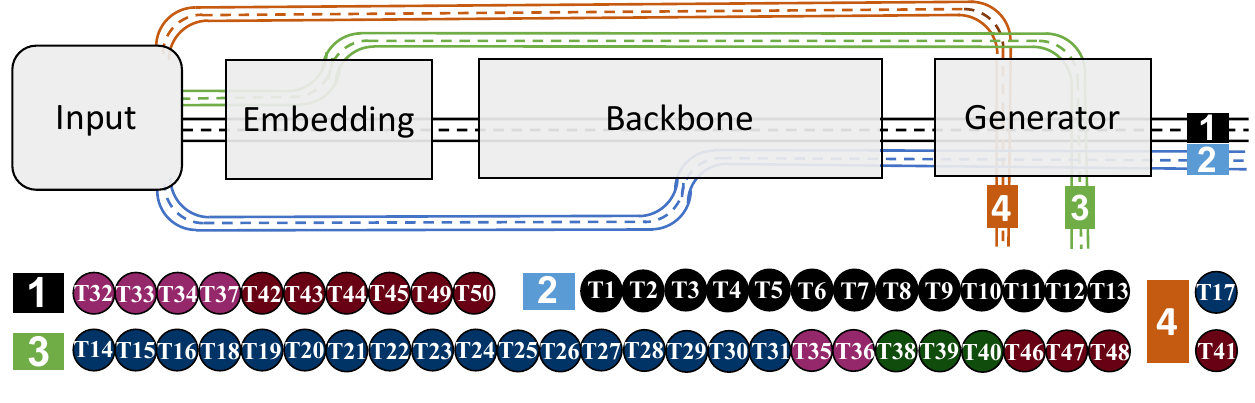}
	\caption{Execution path to each task in Table~\ref{tab:pervasive-tasks}.
	}
	\label{fig:MobileFM-config}
	\vspace{-10pt}
\end{figure}

\paragraph{Task-specific partial activation of \sys.}
Not every task needs to go through the end-to-end workflow of \sys, i.e., embedding-backbone-generator.
Inspired by the early-exit inference~\cite{zhou2020bert, teerapittayanon2016branchynet, laskaridis2021adaptive} and multi-path design in hardware~\cite{wang2022quantumnas,wang2021spatten}, we propose a multi-path task execution design for \sys.
For simpler tasks that can be well solved by only part of \sys's modules, we allow partial activation of \sys to reduce the computing complexity.
In practice, developers could assign specific execution path to different tasks to achieve the best performance.
In Figure~\ref{fig:MobileFM-config}, we have pre-defined 4 paths that can suffice the  50 mobile tasks we have investigated.

\begin{itemize}[leftmargin=*,topsep=1pt]
\item \textbf{Path-1} means a full-model activation of \sys. Tasks taking this path often require cross-modality alignment and complicated task compresion/reasoning. For instance: spoken language understanding and visual question answering.

\item \textbf{Path-2} activates only the backbone and generator, mostly used by language tasks since \sys's backbone can directly take raw textual input.
For instance: input word prediction and machine translation.

\item \textbf{Path-3} activates only the multimodal embedding, often used by tasks that can be accomplished through cross-modality (often language-X) alignment.
For instance, image classification aligns the images with their corresponding textual labels (e.g., ``cat''); human activity recognition tasks align the IMU data with the textual activity types (e.g., ``walking'').
In practice, the labels are embedded into a carefully designed prompt as shown below.

\item \textbf{Path-4} activates only a specific generator, as taken by very few tasks.
For instance: super-resolution can be accomplished by the visual generator (\textit{IMG\_dec}); text-to-speech is accomplished by the speech generator (\textit{TTS\_dec}).
\end{itemize}

\paragraph{Training details}
Fine-tuning \sys follows three steps:
(1) \textit{input data processing:} NLP tasks take as input the context information along with task-specific descriptive language prompts.
CV/Audio/IMU tasks will do alike according to previous work~\cite{girdhar2023imagebind}.
For more intricate tasks, such as object detection, we utilize a region proposal network to generate a set of object proposals~\cite{zhou2022detecting}. 
(2) \textit{tunable weights setting:}
For tasks going through backbone (Path 1, 2), only PEFT parameters in backbones are activated while freezing encoder PEFT modules.
For discriminative tasks that only go through encoders (Path 3), the encoder PEFT modules will be activated.
Linear mapping contacting encoders and backbones will be always activated for shaping alignment.
(3) \textit{model weights updating:}
During each training iteration, we compute the CrossEntropy loss from actual labels and predicted tokens, utilizing it to update the PEFT/MLP parameters.

\begin{table}[]
	\centering
	\resizebox{0.98\columnwidth}{!}{%
	\begin{tabular}{|c|c|l|}
		\hline
		\textbf{Tasks}                 & \textbf{Path} & \multicolumn{1}{c|}{\textbf{Prompts at Text Embedding and Backbone}}                                                                           \\ \hline
		Image classification           & Path-3        & {[}E{]}: There is a photo of a {[}Image\_label: car{]}.                                                                                        \\ \hline
		Machine translation            & Path-2        & \begin{tabular}[c]{@{}l@{}}{[}B{]}: Translate the following sentences from \\ {[}SRC\_language: en{]} to {[}TGT\_language: de{]}.\end{tabular} \\ \hline
		Code generation                & Path-2        & \begin{tabular}[c]{@{}l@{}}{[}B{]}: Write an assembly code according \\ to the {[}sentence{]} requirements.\end{tabular}                       \\ \hline
		HAR     & Path-3        & {[}E{]}: The human is {[}Activity\_label: sitting{]}.                                                                                          \\ \hline
		Audio captioning               & Path-1        & \begin{tabular}[c]{@{}l@{}}{[}E{]}: Give a very short caption of the audio, \\ the caption have 16 words at most.\end{tabular}                 \\ \hline
		Image captioning               & Path-1        & \begin{tabular}[c]{@{}l@{}}{[}E{]}: Give a very short caption of the image, \\ the caption have 16 words at most.\end{tabular}                 \\ \hline
		Video classification           & Path-3        & {[}E{]}: There is a video of {[}Video\_label: abseiling{]}.                                                                                    \\ \hline
		OCR & Path-3        & {[}E{]}: A {[}negative / positive{]} review of a movie.                                                                                        \\ \hline
		\end{tabular}
	}
	\caption{A few prompt examples used in \sys. [E] denotes the Text Embedding. [B] denotes the Backbone.}
	\label{tab:prompts}
	\vspace{-20pt}
\end{table}

\begin{table*}[t]
\centering
\resizebox{2.08\columnwidth}{!}{%
\begin{tabular}{|l|l|l|l|l|l|l|}
\hline
\multirow{2}{*}{\textbf{Category}} & \multirow{2}{*}{\textbf{Tasks}}                & \multirow{2}{*}{\textbf{Mobile Application}}                 & \multirow{2}{*}{\textbf{Dataset}} & \multirow{2}{*}{\textbf{Specific-Models}} & \multirow{2}{*}{\textbf{Results}} & \multirow{2}{*}{\textbf{Metrics}} \\
                                   &                                                &                                                              &                                   &                                           &                                    &                                    \\ \hline
\multirow{13}{*}{NLP}              & Input word prediction   \circleA{\color0}{T1}                           & Input method (GBoard)                                        & PTB                               & RNN \cite{Bench1}                                      & 0.17*                          & Accuracy                              \\ \cline{2-7} 
                                   & \multirow{2}{*}{Question answering \circleA{\color0}{T2}    \circleA{\color0}{T3}    }            & \multirow{2}{*}{Private assistant (Siri)}                    & SQuAD v2.0                        & RoBERTa \cite{Bench2}                                  & 0.79*                        & F1                               \\ \cline{4-7} 
                                   &                                                &                                                              & TyDi QA                           & AraELECTRA  \cite{Bench3}                              & 0.87                          & F1                               \\ \cline{2-7} 
                                   & Machine translation  \circleA{\color0}{T4}                              & Translator (Google Translate)                                & wmt22 en-de                       & Transformer \cite{Bench4}                               & 0.34*                       & BLEU                               \\ \cline{2-7} 
                                   & Emoji prediction   \circleA{\color0}{T5}                                & Input method (GBoard)                                        & tweet\_eval               & RoBERTa \cite{Bench5}                                  & 0.33*                         & Accuracy                               \\ \cline{2-7} 
                                   & Emotion prediction  \circleA{\color0}{T6}                               & Conversational analytics (Clarabridge)                       & go\_emotion                       & RoBERTa  \cite{Bench6}                                   & 0.57*                        & Accuracy                               \\ \cline{2-7} 
                                   & Sentiment analysis  \circleA{\color0}{T7}                               & Conversational analytics (Clarabridge)                       & tweet\_eval                       & RoBERTa  \cite{Bench7}                                 & 0.77*                          & Accuracy                               \\ \cline{2-7} 
                                   & \multirow{2}{*}{Text classification \circleA{\color0}{T8}    \circleA{\color0}{T9}    }           & \multirow{2}{*}{Spam SMS filtering (Truecaller)}             & ag\_news                          & BERT \cite{Bench8}                                      & 0.93*                        & Accuracy                               \\ \cline{4-7} 
                                   &                                                &                                                              & SST2                              & DistilBERT  \cite{Bench9}                              & 0.91*                        & Accuracy                               \\ \cline{2-7} 
                                   & Grammatical error correction  \circleA{\color0}{T10}                     & Writing assistant (Grammarly)                                & JFLEG                             & FLAN-t5 \cite{Bench10}                                  & 0.68*                        & BLEU                               \\ \cline{2-7} 
                                   & Text summary  \circleA{\color0}{T11}                                     & Reading assistant (ChatPDF)                                  & CNN Daily Mail                    & BART \cite{Bench11}                                     & 0.43*                     & ROUGE1                               \\ \cline{2-7} 
                                   & Code document generation  \circleA{\color0}{T12}                         & Code editor (Javadoc)                                        & CodeSearchNet                     & CodeT5-base \cite{Bench12}                              & 0.33*                     & ROUGE1                               \\ \cline{2-7} 
                                   & Code generation   \circleA{\color0}{T13}                                 & Code editor (Copilot)                                        & Shellcode\_IA32                   & CodeBERT \cite{Bench13}                                 & 0.92                        & BLEU                               \\ \hline
\multirow{18}{*}{CV}               & \multirow{2}{*}{Object detection \circleD{\color3}{T14} \circleD{\color3}{T15}}              & \multirow{2}{*}{Augmented Reality (Google Lens)}             & COCO                              & Libra-rcnn \cite{Bench14}                                & 0.43*                       & mAP                               \\ \cline{4-7} 
                                   &                                                &                                                              & LVIS                              & X-Paste  \cite{Bench15}                                 & 0.51                          & AP                               \\ \cline{2-7} 
                                   & Image retrieval \circleD{\color3}{T16}                               & Image searcher (Google Photos)                               & Clothes Retrieval                 & Resnet50-arcface \cite{Bench16}                         & 0.90*                      & Recall                               \\ \cline{2-7} 
                                   & Super-resolution \circleD{\color3}{T17}                              & Video/Image super-resolution (VSCO)                          & set5                              & Real-ESRGAN \cite{Bench17}                              & 0.82*                      & SSIM                               \\ \cline{2-7} 
                                   & Styler transfer  \circleD{\color3}{T18}                              & Painting \& Beatifying (Meitu)                               & COCO, Wikiart                     & StyleGAN-nada \cite{Bench18}            & 0.23                        & CLIP score                               \\ \cline{2-7} 
                                   & \multirow{2}{*}{Semantic segmentation \circleD{\color3}{T19} \circleD{\color3}{T20}}         & \multirow{2}{*}{Smart camera (Segmentix)}                    & ADE20K-150                        & Deeplabv3plus \cite{Bench19}                            & 0.43*                      & mIoU                               \\ \cline{4-7} 
                                   &                                                &                                                              & PASCAL VOC                   & Deeplabv3plus \cite{Bench20}                            & 0.79*                       & mIoU                               \\ \cline{2-7} 
                                   & Optical character recognition  \circleD{\color3}{T21}                & \begin{tabular}[c]{@{}l@{}} Intelligent document automation \\ software (Ocrolus) \end{tabular}          & Rendered SST2                     & CLIP \cite{Bench21_27_47}                                      & 0.71                             & Accuracy                            \\ \cline{2-7} 
                                   & \multirow{2}{*}{Image classification \circleD{\color3}{T22} \circleD{\color3}{T23}}          & \multirow{2}{*}{Album management (Google Photos)}            & CIFAR100                          & GFNet-XS  \cite{Bench22}                                & 0.89                          & Accuracy                               \\ \cline{4-7} 
                                   &                                                &                                                              & ImageNet                          & Resnet-152 \cite{Bench23}                               & 0.79                          & Accuracy                               \\ \cline{2-7} 
                                   & Traffic sign classification \circleD{\color3}{T24}                   & Intelligent transportation (Waze)                            & GTSRB                             & MicronNet \cite{Bench24}                                & 0.98                         & Accuracy                               \\ \cline{2-7} 
                                   & Vehicle re-identification  \circleD{\color3}{T25}                    & Surveillance camera (AI Re-ID)                               & Veri776                           & MSINet \cite{Bench25}                                   & 0.96                         & Rank                               \\ \cline{2-7} 
                                   & Gender recognition  \circleD{\color3}{T26}                           & Smart camera (Face++)                                        & Adience                           & MiVOLO-D1 \cite{Bench26}                                & 0.96                          & Accuracy                               \\ \cline{2-7} 
                                   & Location recognition  \circleD{\color3}{T27}                         & Navigation search (Google Maps)                              & Country211                        & CLIP  \cite{Bench21_27_47}                                    & 0.46                         & Accuracy                               \\ \cline{2-7} 
                                   & Pose estimation  \circleD{\color3}{T28}                              & AI fitness coach (Keep)                                      & AP-10K                            & ViTPose \cite{Bench28}                                   & 0.69                          & AP                               \\ \cline{2-7} 
                                   & Video classification   \circleD{\color3}{T29}                        & Video player (YouTube)                                       & kinetics400                       & SlowFast  \cite{Bench29}                                 & 0.79                          & Accuracy                               \\ \cline{2-7} 
                                   & Crowd Counting  \circleD{\color3}{T30}                               & Smart camera (Fitness Tracking)                              & UCF-QNRF                          & CSS-CCNN  \cite{Bench30}                                 & 437                           & MAE                               \\ \cline{2-7} 
                                   & Image matting  \circleD{\color3}{T31}                                & Virtual backgrounds (Zoom)                                   & RefMatte-RW100                    & MDETR     \cite{Bench31}                                 & 0.06                         & MSE                               \\ \hline
\multirow{6}{*}{Audio}             & Automatic speech recognition   \circleE{\color4}{T32}                 & Private assistant (Siri)                                     & LibriSpeech                       & CTC+attention \cite{Bench32}                             & 3.16\%*                                  & WER                               \\ \cline{2-7} 
                                   & \multirow{2}{*}{Spoken language understanding \circleE{\color4}{T33}  \circleE{\color4}{T34} } & \multirow{2}{*}{Private assistant (Siri)}                    & FSC                               & Transformer \cite{Bench33}                               & 0.37\%                                  & WER                               \\ \cline{4-7} 
                                   &                                                &                                                              & SLURP                             & CRDNN \cite{Bench34}                                     & 0.82*                          & Accuracy                               \\ \cline{2-7} 
                                   & Emotion recognition \circleE{\color4}{T35}                            & Emoji recommendation (WeChat)                                & IEMOCAP                           & ECAPA-TDNN  \cite{Bench35}                               & 0.64*                         & Accuracy                               \\ \cline{2-7} 
                                   & Audio classification  \circleE{\color4}{T36}                          & Music discovery (Shazam)                                     & ESC-50                            & ACDNet  \cite{Bench36}                                 & 0.87                          & Accuracy                               \\ \cline{2-7} 
                                   & Keyword spotting   \circleE{\color4}{T37}                             & Private assistant (Siri)                                     & Speech command                    & Cnn-trad-fpool3  \cite{Bench37}                          & 0.88*                        & Accuracy                               \\ \hline
\multirow{3}{*}{Sensing}           & \multirow{3}{*}{Human activity recognition \circleB{\color1}{T38} \circleB{\color1}{T39} \circleB{\color1}{T40} }    & \multirow{3}{*}{AI fitness coach (Keep)}                     & Using Smartphones                 & TS-TCC \cite{Bench38}                                    & 0.90                          & Accuracy                               \\ \cline{4-7} 
                                   &                                                &                                                              & HHAR                              & LIMU-BERT  \cite{Bench39_40}                                & 0.84                          & Accuracy                               \\ \cline{4-7} 
                                   &                                                &                                                              & MotionSense                       & LIMU-BERT  \cite{Bench39_40}                               & 0.91                          & Accuracy                               \\ \hline
\multirow{10}{*}{Multimodal}       & Text-to-speech \circleC{\color2}{T41}                                & Voice broadcast (WeChat reading)                             & LJSpeech                          & Transformer \cite{Bench41}                               & 3.26                          & MCD                               \\ \cline{2-7} 
                                   & \multirow{2}{*}{Audio captioning \circleC{\color2}{T42} \circleC{\color2}{T43}}              & \multirow{2}{*}{Hearing-impaired accessibility (Ava)}        & Clotho                            & Transformer \cite{Bench42_43}                               & 0.52*                       & BLEU                               \\ \cline{4-7} 
                                   &                                                &                                                              & AudioSet                          & Transformer  \cite{Bench42_43}                             & 0.64                        & BLEU                               \\ \cline{2-7} 
                                   & \multirow{2}{*}{Image captioning \circleC{\color2}{T44} \circleC{\color2}{T45}}              & \multirow{2}{*}{\begin{tabular}[c]{@{}l@{}}Visual-impaired accessibility\\  (Supersence)\end{tabular}}  & MSCOCO'14                         & LSTM \cite{Bench44}                                        & 0.73*                         & BLEU                               \\ \cline{4-7} 
                                   &                                                &                                                              & Flickr8k                          & LSTM \cite{Bench45}                                      & 0.58                        & BLEU                               \\ \cline{2-7} 
                                   & \multirow{2}{*}{Text-to-image retrieval \circleC{\color2}{T46} \circleC{\color2}{T47}}       & \multirow{2}{*}{Image search (Google Photos)}                & Flickr8k                          & NAPReg \cite{Bench46}                                      & 0.39                      & Recall                               \\ \cline{4-7} 
                                   &                                                &                                                              & Flickr30k                         & CLIP \cite{Bench21_27_47}                                     & 0.69                       & Recall                               \\ \cline{2-7} 
                                   & Audio/Text-to-image generation  \circleC{\color2}{T48}               & Art creation (Verb Art)                                      & VGGSound                          & Wav2clip \cite{Bench48}                               & 99.89                        & FID                               \\ \cline{2-7} 
                                   & \multirow{2}{*}{Visual question answering \circleC{\color2}{T49} \circleC{\color2}{T50}}     & \multirow{2}{*}{\begin{tabular}[c]{@{}l@{}}Visual-impaired accessibility\\  (Answerables)\end{tabular}} & VQA v2.0                          & MUTAN \cite{Bench49_50}                                     & 0.63                         & Accuracy                               \\ \cline{4-7} 
                                   &                                                &                                                              & VizWiz                            & MUTAN \cite{Bench49_50}                                     & 0.52                        & Accuracy                               \\ \hline
\end{tabular}
}
\caption{A comprehensive benchmark of edge-oriented AI tasks. Circled abbreviation denotes specific task and dataset. 
* denotes the results obtained from Jetson ORIN, while others are obtained from A100.
}
\label{tab:pervasive-tasks}
\vspace{-15pt}
\end{table*}

\paragraph{Prompt design}
Two parts of \sys need careful prompt engineering \cite{jiang2020can, shin2020autoprompt, zhou2022large} to fully exploit its potential: the text embedding and foundation backbone.
We have designed prompts for each mobile AI task in $\S$\ref{subsec:eval-setup}, and Table~\ref{tab:prompts} lists a few of them as exemplifications.

    \section{Experiments and Analysis} 
\label{sec:exps}

\subsection{Benchmark and Setups}
\label{subsec:eval-setup}



\noindent \textbf{\benchmark: a comprehensive edge-oriented benchmark for AI tasks.}
As the very first effort for a one-size-fits-all mobile foundation model, a pivotal undertaking is the comprehensive evaluation of its versatility across diverse mobile AI tasks.
Therefore, we embark on constructing an exhaustive 
edge-oriented benchmark for AI tasks,
encompassing 38 tasks spanning 50 public datasets, as shown in Table~\ref{tab:pervasive-tasks}.
Those tasks are essential to real-world mobile applications (e.g., translation, object detection, and voice assistant). 
Many of these tasks have also received substantial attention within the mobile community itself~\cite{kong2023convrelu++, jin2023emsassist, park2023omnilive, cao2019deqa, yuan2022infi, jiang2021flexible, shi2021face, zhang2021elf, wang2020you, xu2018deepcache}.
Each task is accompanied by its designated accuracy metric.
\benchmark includes 5 modality domains: NLP, CV, Audio, Sensing (IMU), and Misc (Multimodal).
While the majority of tasks are tailored to smartphones, we extend our scope to encompass pivotal devices such as laptops (code generation), autonomous cars (traffic sign classification), and IoT cameras (Counting).

To understand \sys's performance, we select one task-specific model (namely \texttt{TS-model}) for each dataset as a baseline.
The selection of these models adheres to two primary criteria:
(1) They must remain within the confines of mobile device resource constraints, specifically with fewer than 1 billion parameters;
(2) The model accuracy shall be representative to the status quo on mobile devices, though not necessary to be absolute state-of-the-art.
Consequently, our model selection draws primarily from two sources: open-source endeavors showcased at prominent mobile conferences like MobiCom and MobiSys during the past five years, and contemporary models showcased on the Paperwithcode platform \cite{Paperwithcode}.
A comprehensive list of employed \texttt{TS-models} is provided in Table~\ref{tab:pervasive-tasks}, including instances such as ResNet-152 \cite{he2016deep} for image classification, RoBERTa \cite{liu2019roberta} for question answering, and CRDNN \cite{xiang2020optimization} for spoken language understanding.

\noindent \textbf{Hardware}
We use three kinds of hardware platform: 
\begin{itemize}[leftmargin=*]
    \item NVIDIA A100 GPU (A100), a high performance accelerator released in May. 2020 that has 40GB RAM, 384GB storage.
    \item NVIDIA Jetson ORIN NX (Orin), a high-end edge board for autonomous robotics or cars released in Feb. 2023 that has 16GB RAM, 64GB storage, and 1024-core NVIDIA Ampere architecture GPU with 32 Tensor Cores.
    \item Pixel 7 Pro smartphone (Pixel), a high-end smartphone released in Oct. 2022 that has 12GB RAM, 256GB storage, Octa-core CPU, Mali-G710 MP7 GPU, and an edge TPU.
\end{itemize}

All training experiments are conducted on two GPU servers, each equipped with 8 A100s. 
This encompassed the fine-tuning of \sys and a portion of the training for the TS-model starting from scratch.
To facilitate the accuracy of \texttt{TS-models} on each dataset quickly, we performe the inference experiments on A100 and Orin.
We show the comprehensive results and corresponding platforms in Table~\ref{tab:pervasive-tasks}.
In total, our experiments take 100,000 GPU-hours.
We use ORIN and Pixel, two typical mobile devices, to measure the runtime performance (memory, latency, energy, etc.) of \sys on real-world devices.




\noindent \textbf{Implementation}
The benchmark results are tested with PyTorch~\cite{pytorch}, and model parameters were obtained through two methods:
(1) Directly downloading pre-trained parameters from open-source websites, such as Hugging Face~\cite{Huggingface};
(2) Training model parameters from scratch based on the open-source code, available on platforms like GitHub~\cite{github}.


The \sys prototype is built upon PandaGPT~\cite{su2023pandagpt}, a versatile instruction-following foundation model. Additionally, we implement two crucial modules using PyTorch~\cite{pytorch}:
(1) A redesigned multimodal generator to broaden output capabilities for mobile scenarios, including classification, text-to-speech, and image generation, surpassing the exclusive focus on text generation.
(2) A multi-path controller aimed at enhancing compatibility with diverse mobile AI tasks.
We directly acquire pre-trained parameters for the embedding and backbone from official releases~\cite{git-llama,git-imagebind}. 
Following previous researches~\cite{hu2021lora,li2023loftq}, we fine-tune their adapters from scratch, as elaborated in Section \ref{subsec:peft}. 

The runtime-cost performance on Orin is obtained through jetson-stats~\cite{jetson-stats}, which is a powerful tool to analyze your board.
Regarding the Pixel smartphone, the latency and memory results are measured using TFLite's benchmark tools~\cite{tflitetools}, while power consumption data is extracted from Android's virtual file system (e.g., /sys and /proc). 

\begin{figure*}[t]
	\centering
	\includegraphics[width=1\textwidth]{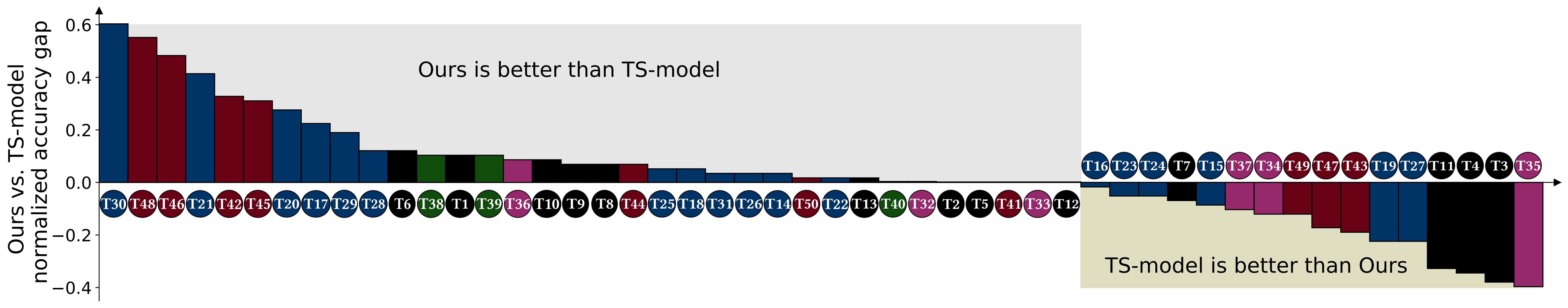}
	\caption{Normalized accuracy comparison of \sys and \texttt{TS-models} on 50 popular mobile tasks and datasets.}
	\label{fig:eval-acc-gap}
	\vspace{-5pt}
\end{figure*}
\subsection{Overall Accuracy}
\label{subsec:accuracy}







\begin{figure}[t]
	\centering
	\includegraphics[width=0.48\textwidth]{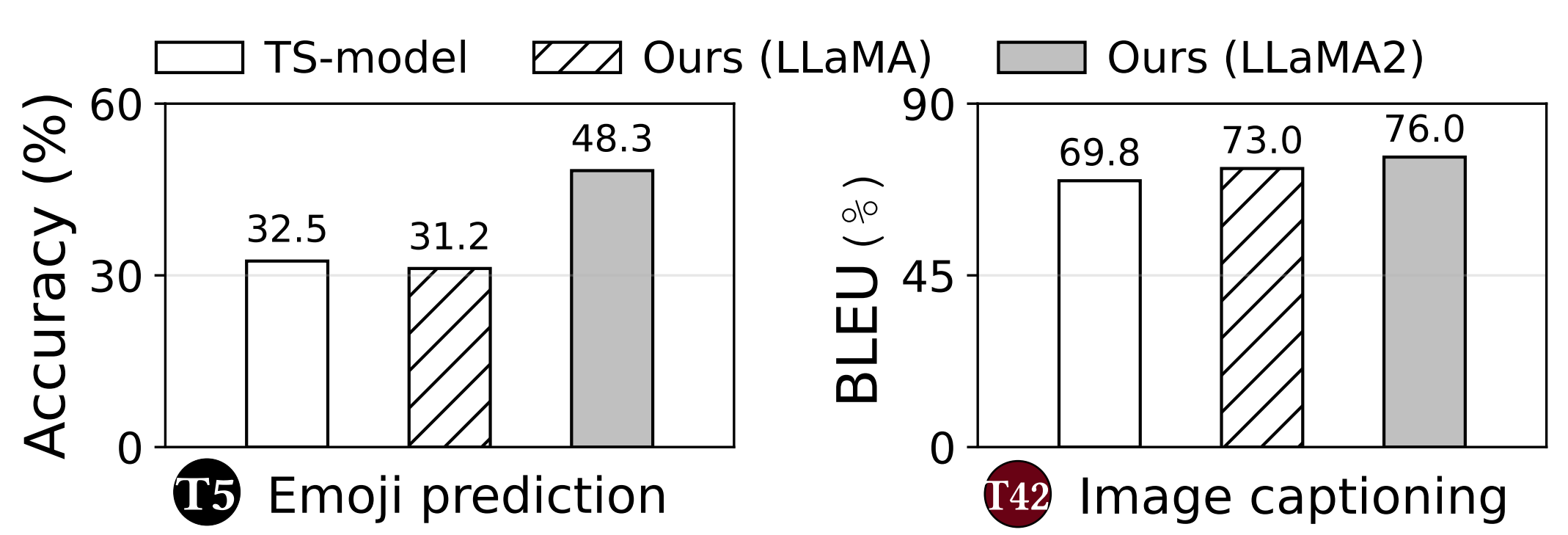}
	\caption{Performance improvement of \sys when replacing LLaMA with LLaMA2 as foundation backbone.}
	\label{fig:eval-acc-llama2}
	\vspace{-15pt}
\end{figure}

\begin{figure}[t]
	\centering
	\includegraphics[width=0.48\textwidth]{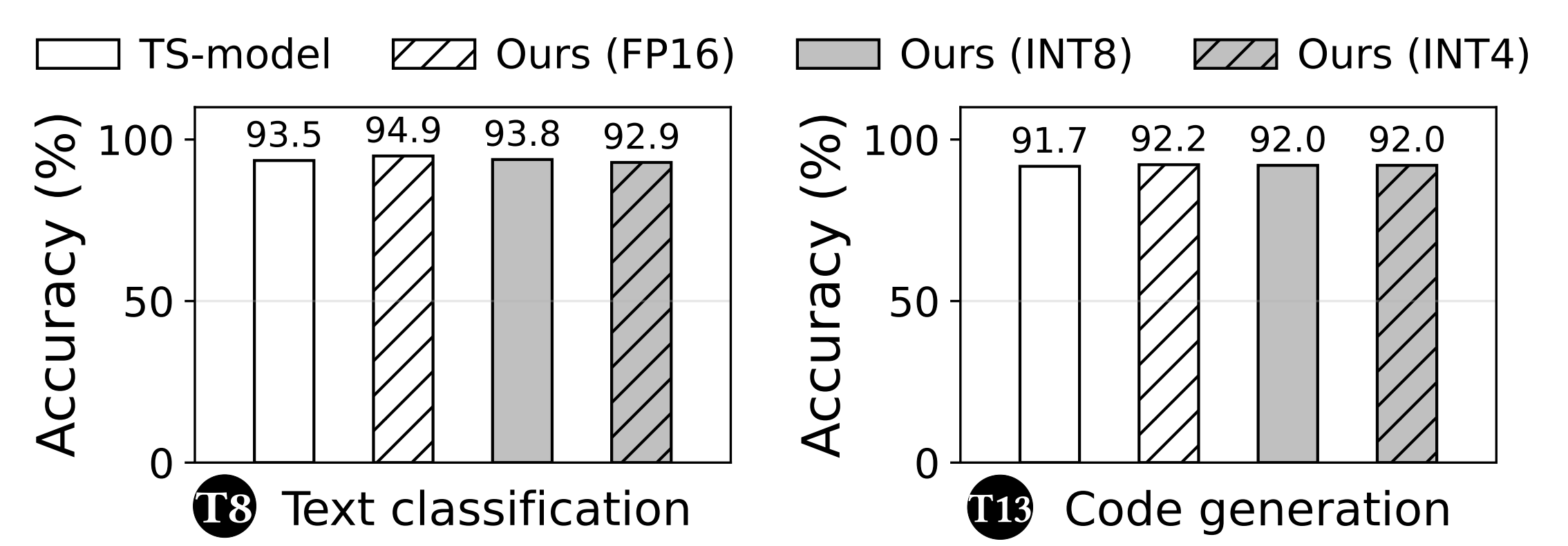}
	\caption{Accuracy of quantized \sys compared with \texttt{TS-models}. FP16/INT8/INT4 represent the numerical representation bit-width employed by LLaMA. }
	\label{fig:eval-acc-quantization}
	\vspace{-15pt}
\end{figure}

\noindent \textbf{\sys can well support most mobile AI tasks and datasets.}
Figure~\ref{fig:eval-acc-gap} illustrates \sys's overall performance improvement (or degradation) compared to \texttt{TS-models}.
As observed, \sys can achieve comparable performance across 85\% of tasks, with over 50\% of these tasks showcasing considerable performance improvement.
Note that the vertical axis, normalized in Table 3, reflects variations in accuracy across the 50 tasks. Our main focus is on evaluating whether M4 demonstrates superior or inferior accuracy compared to respective tasks, assessing its universal capability. Emphasis is placed on overall performance rather than specific task improvements.
For example on \circleAS{\color0}{T1}input word prediction, \circleCS{\color2}{T42}audio captioning, and \circleCS{\color2}{T46}text-to-image retrieval, \sys yields accuracy increments of 6\%, 19\%, and 28\% respectively.
Such commendable improvements are attributed to the well-engineered design of \sys, characterized by its unified, adaptive, and multimodal foundation model.
While \sys's prowess is manifest, it is prudent to acknowledge marginal performance dips (not surpassing 10\%) observed in specific tasks. Instances such as \circleAS{\color0}{T7} sentiment analysis, \circleBS{\color1}{T16} image retrieval, and \circleES{\color4}{T37} keyword spotting exemplify this trend, with accuracy experiencing nominal decrements of 4\%, 1\%, and 6\%, respectively. 
However, even in these cases, \sys remains viable for deployment with usable performance.
Additionally, \sys only showcases diminished performance in 4 tasks, with accuracy drop-offs of up to 20\%. 
The reason behind this reduction stems from the unique requirements of certain low-resource translation tasks, necessitating extensive language knowledge that isn't inherently embedded within the current foundation model's pre-training phase.

\noindent \textbf{\sys can be further enhanced with enhanced foundation models.}
Figure~\ref{fig:eval-acc-llama2} illustrates the performance improvement realized by \sys through the integration of the latest LLaMA2, in comparison to LLaMA.
LLaMA2, a refined evolution of its precursor, LLaMA, introduces heightened capabilities and marked improvements~\cite{touvron2023llama}. Released in July 2023, LLaMA2 marks a substantial leap forward, expanding the context window and ushering in the innovation of grouped-query attention. This novel architectural element empowers the model with rapid information processing capabilities.
As observed, \sys using LLaMA2 attains a remarkable 15\% accuracy enhancement and a 2\% improvement in BLEU scores for \circleAS{\color0}{T5} emoji prediction and \circleCS{\color2}{T42} image caption tasks, respectively. 
This prowess is attributed to LLaMA2's optimized architectural schema, expansive training corpus (comprising 2T tokens), and elevated data quality~\cite{touvron2023llama}.
As \sys is inherently adaptable to its foundation underpinnings, it seamlessly integrates and capitalizes upon the latest components. 

\noindent \textbf{\sys can efficiently preserve the performance with low-bit quantization.}
Figure~\ref{fig:eval-acc-quantization} illustrates the performance comparison of \sys using quantized backbone with respect to the \texttt{TS-model}.
As observed, \sys using 8-bit (INT8) and 4-bit (INT4) quantization both achieve nearly lossless accuracy, compared to \sys using 16-bit float representation (FP16).
For example, on \circleAS{\color0}{T8}text classification and \circleAS{\color0}{T13}code generation, INT8 and INT4-based M4 achieved only a marginal decrease in accuracy compared to FP16-based \sys, with reductions of 0.2\%-0.9\% and 0.2\%-2\%, respectively.
The reasonable behind is that large models possess an abundance or even surplus knowledge representation, which contributes to more extensive knowledge even after quantization~\cite{frantar2022gptq}.
Therefore, we consistently employed \sys with the default INT8 quantization of the LLaMA backbone.



\subsection{Zero/Few-shot Ability}
\label{subsec:fewshot}


\begin{figure}[t]
	\begin{subfigure}{.235\textwidth}
		\centering
		\includegraphics[width=\textwidth]{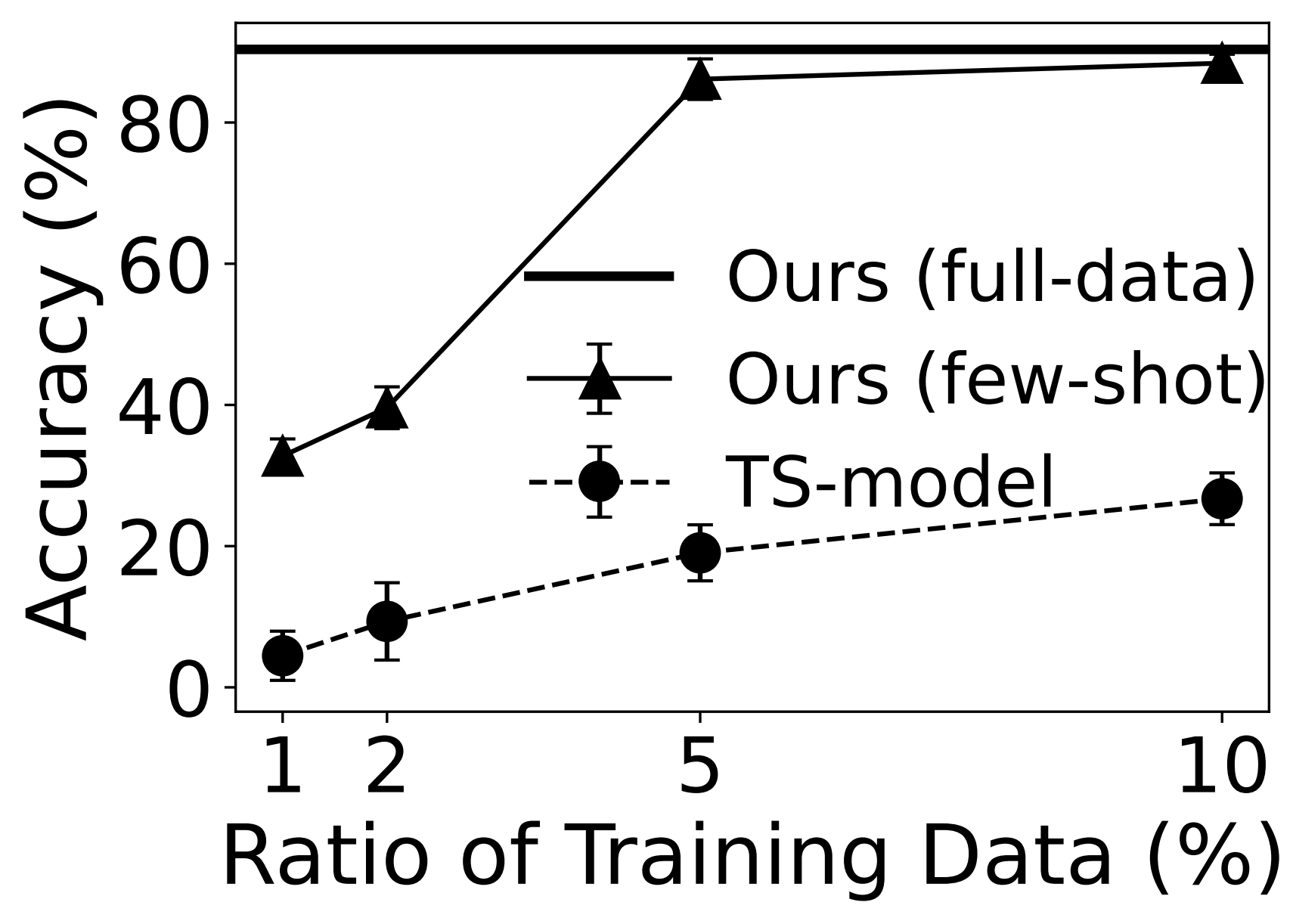}
		\caption{
		Image classification}
	\end{subfigure}
	\begin{subfigure}{.235\textwidth}
		\centering
		\includegraphics[width=\textwidth]{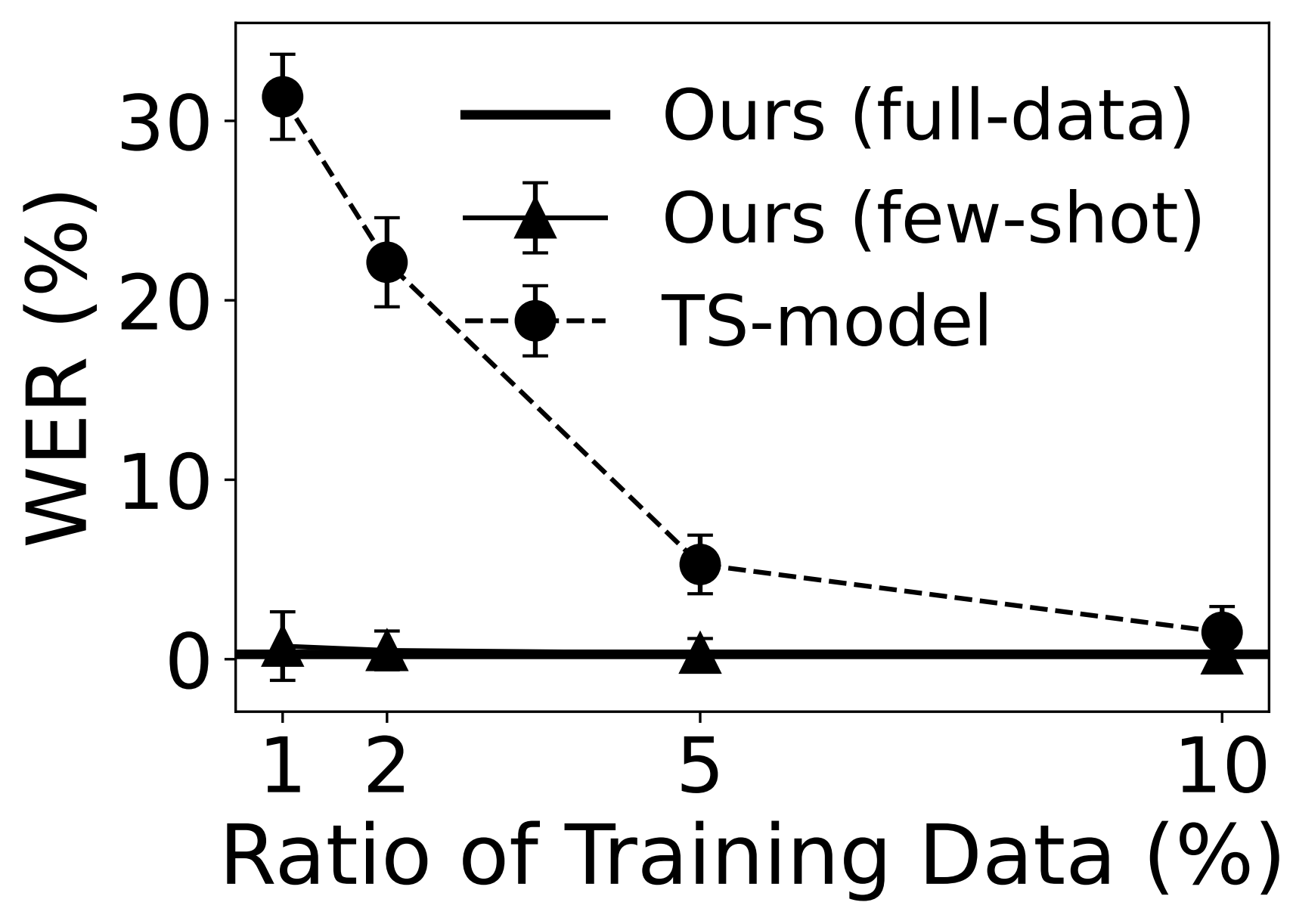}
		\caption{
			SLU (lower is better)}
	\end{subfigure}
	\caption{Few-shot testing of \sys and \texttt{TS-model}. SLU: spoken language understanding.
		}
	\label{fig:few-shot}
	\vspace{-12pt}
\end{figure}
We experiment on two tasks, image classification and spoken language understanding.
For each task, we follow prior work \cite{cai2022fedadapter} to randomly select gold labels, with the sample size varying between 1\% and 10\% of the entire dataset.
By default, the labels form a skewed distribution across clients to be more realistic to real-world situations.
For each dataset, we conduct 5 repeated experiments and report the mean results.

\noindent \textbf{\sys has better few-shot ability than \texttt{TS-models} that are trained from scratch.}
In Figure~\ref{fig:few-shot}, few-shot \sys performs on par with or slightly lower than \sys with full data tuning. 
Notably, it consistently outperforms \texttt{TS-model} by up to 67.1\%.
For example on \circleES{\color4}{T33}, even with a mere 1\% sample (equating to just 231 samples), few-shot \sys achieves a Word Error Rate (WER) of 0.7\%. 
It is a mere 0.4\% higher than \sys with full data, but 25.4\% lower than \texttt{TS-model} that is trained from scratch.  
This outcome underscores \sys's prowess in leveraging pre-trained multimodal knowledge for swift adaptation to new tasks, even with scarce data.

\begin{figure}[t]
	\centering
	\includegraphics[width=0.48\textwidth]{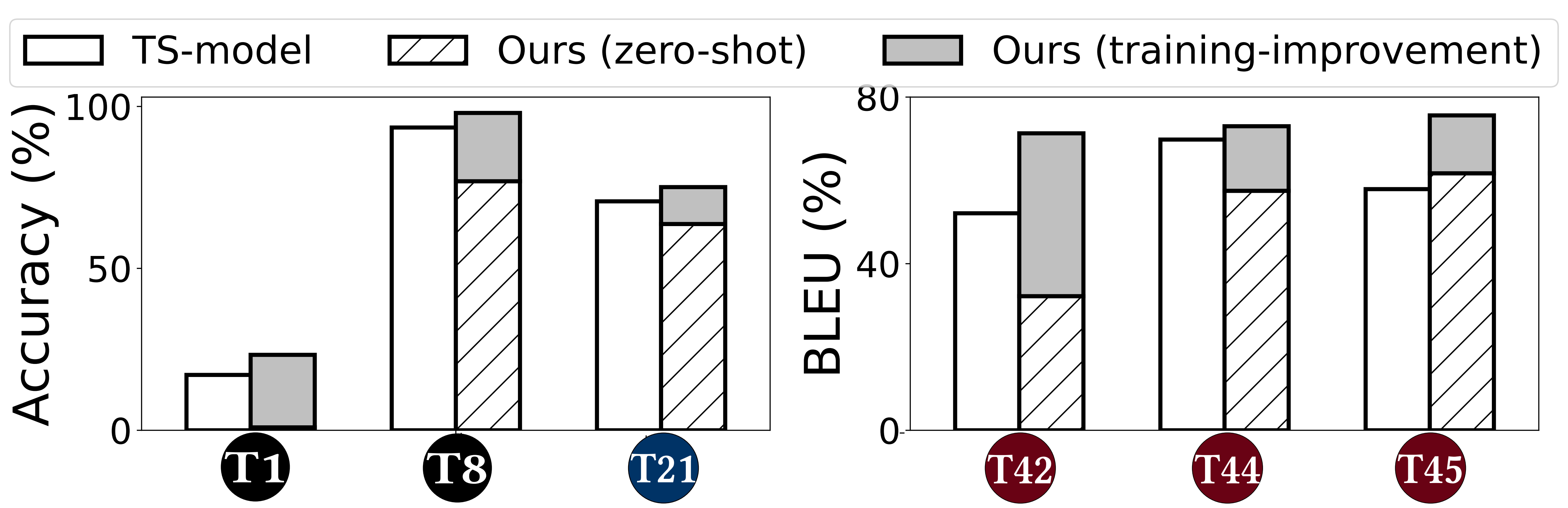}
	\caption{Zero-shot testing of \sys and \texttt{TS-model}.}
	\label{fig:zeroshot}
	\vspace{-10pt}
\end{figure}
\noindent \textbf{\sys also has a certain zero-shot ability, but fine-tuning makes it much more accurate.}
Figure~\ref{fig:zeroshot} illustrates \sys's zero-shot capabilities on 6 tasks.
Evidently, \sys demonstrates commendable zero-shot proficiency, attaining approximately 80\% of the \texttt{TS-model} performance in most cases. Notable instances include \circleAS{\color0}{T8} \circleDS{\color3}{T21} \circleCS{\color2}{T42} and \circleCS{\color2}{T44}, where \sys's zero-shot performance remains acceptably close to the corresponding \texttt{TS-models}, with reductions ranging from 7\% to 20\%.
In \circleCS{\color2}{T45}, \sys showcases a 4\% improvement over \texttt{TS-models}, a testament to the efficacy of prompt learning methodologies. 
Notwithstanding these accomplishments, the application of fine-tuning to these datasets yields substantial accuracy enhancements for \sys, surging by 11\%-39\%. 
This improvement arises from \sys's robust attention-based architecture~\cite{wei2021finetuned}.

\subsection{Parameter-efficient Fine-tuning}
\label{subsec:peft}

\begin{table}[t]
	\centering
	\resizebox{1\columnwidth}{!}{%
	\begin{tabular}{|c|cc|cc|}
		\hline
		\multirow{2}{*}{\textbf{Tasks}} & \multicolumn{2}{c|}{\textbf{PEFT settings}}                                                          & \multicolumn{2}{c|}{\textbf{PEFT results}}                                     \\ \cline{2-5} 
										& \multicolumn{1}{c|}{Techniques} & Rank                                                       & \multicolumn{1}{c|}{ Size (Ratio)} & Acc (Dif)              \\ \hline
		Emoji prediction                          & \multicolumn{1}{c|}{LoRA}     & 4                                                             & \multicolumn{1}{c|}{2M (0.03\%)}       & 31 (1↓)                                 \\ \hline
		Image classification                       & \multicolumn{1}{c|}{LoRA}       & 4                                                             & \multicolumn{1}{c|}{8M (0.007\%)}        & 90 (1↑)                      \\ \hline
		Human activity recognition                         & \multicolumn{1}{c|}{LoRA}       & 1                                                             & \multicolumn{1}{c|}{5M (0.004\%)}        & 96 (5↑)                      \\ \hline
		Audio captioning                        & \multicolumn{1}{c|}{LoRA}       & 4                                                             & \multicolumn{1}{c|}{4M (0.06\%)}          & 72 (19↑) \\ \hline
		\end{tabular}
		
}
\caption{PEFT-enhanced \sys's optimal results. Size (Ratio) denotes the trainable parameter size and its ratio to total parameters. Acc (Dif) denotes the performance of PEFT-enhanced \sys along with the differences compared to \texttt{TS-model}, and the units are \%.}
\label{tab:peft-tabs}
\vspace{-20pt}
\end{table}


\begin{figure}[t]
	\begin{subfigure}{.235\textwidth}
		\centering
		\includegraphics[width=\textwidth]{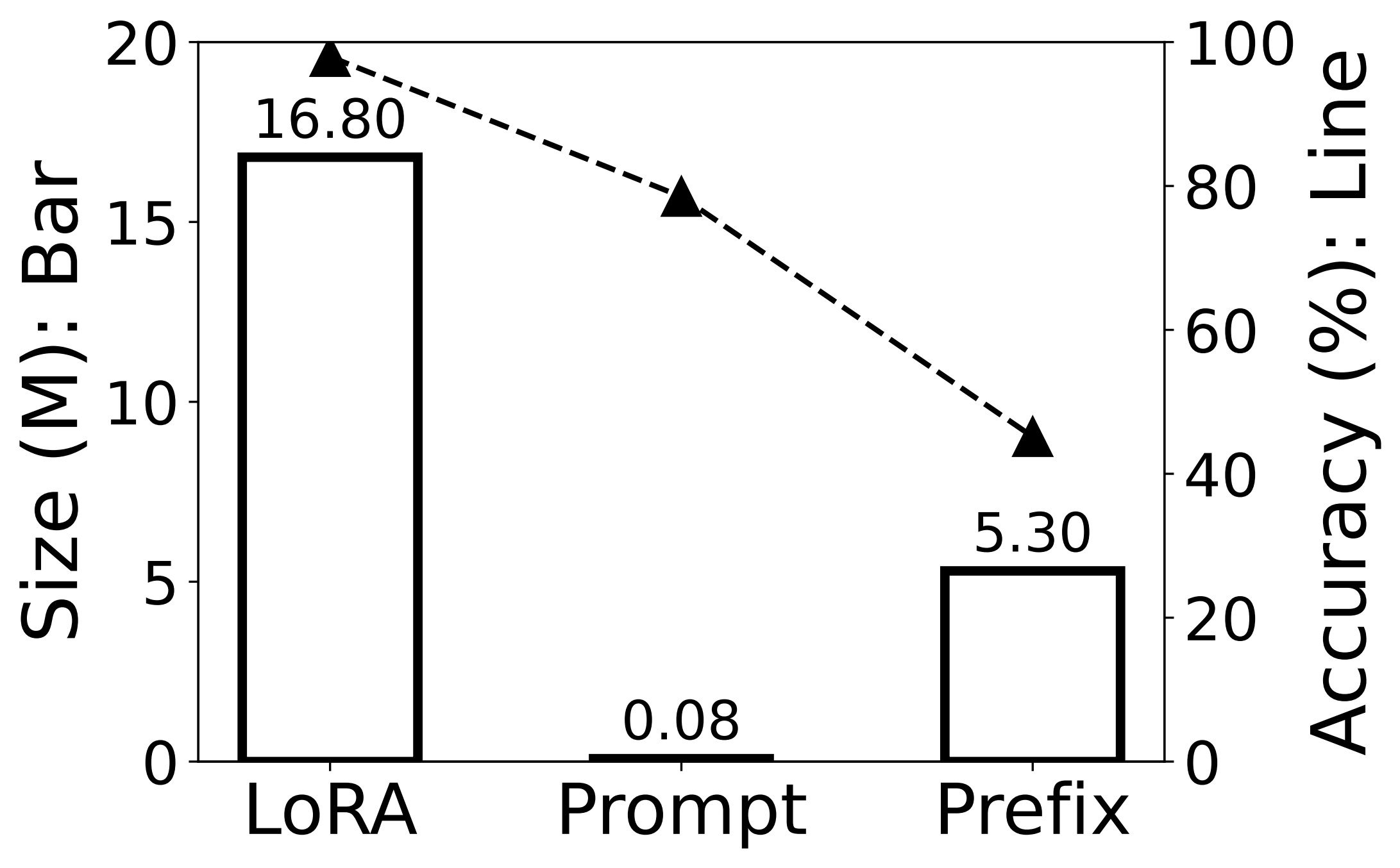}
		\caption{PEFT techniques}
	\end{subfigure}
	\begin{subfigure}{.235\textwidth}
		\centering
		\includegraphics[width=\textwidth]{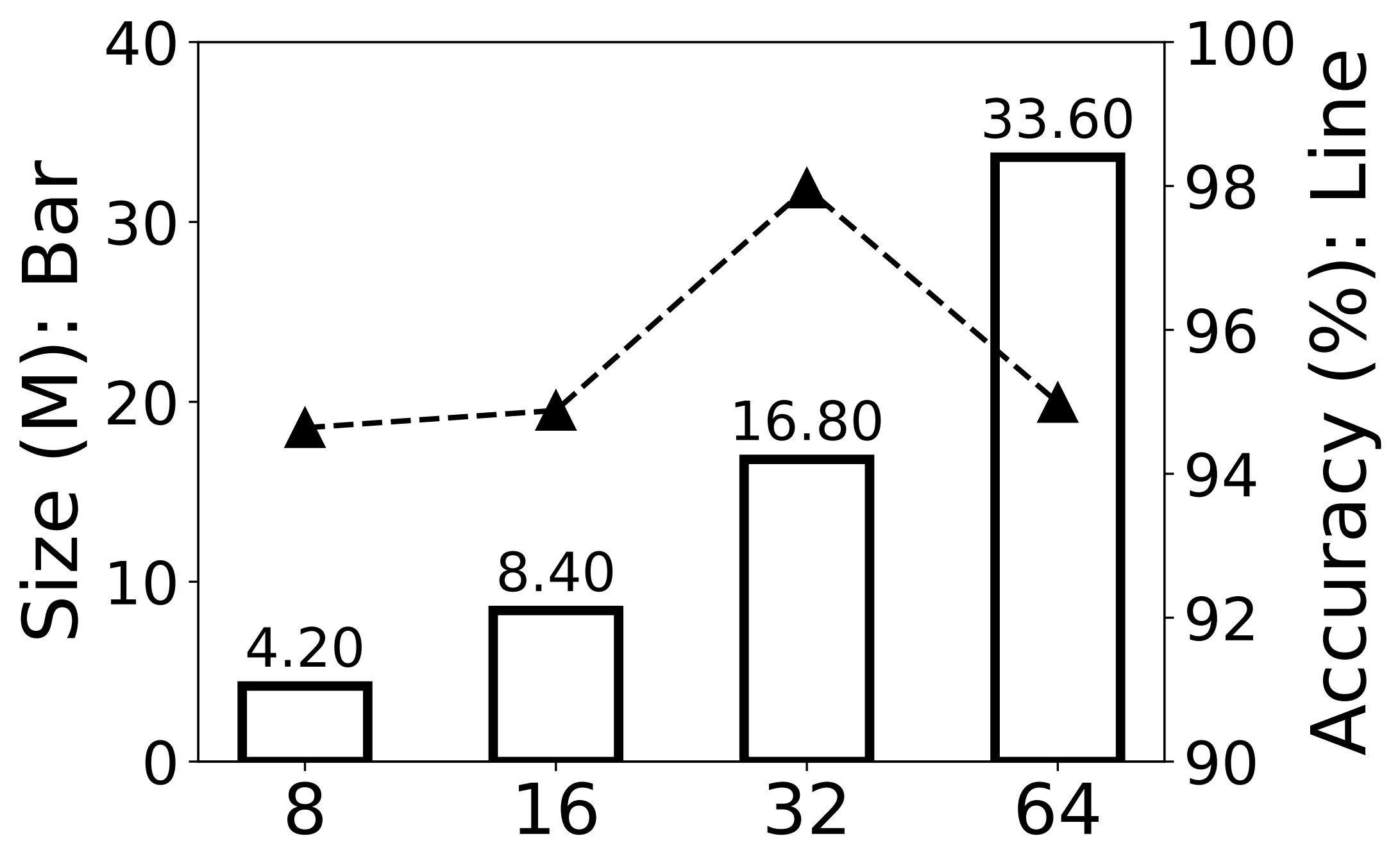}
		\caption{PEFT LoRA rank}
	\end{subfigure}
	\caption{Impact analysis of PEFT-enhanced \sys on text classification. Size (M): trainable parameter size.}
	\label{fig:peft_time}
	\vspace{-10pt}
\end{figure}


\noindent \textbf{A proper PEFT technique and its configuration is crucial to trade off \sys performance and cost.}
Table~\ref{tab:peft-tabs} reports the optimal results of \sys on the trade-off between model accuracy and trainable parameter size. 
Our observations highlight the efficacy of the LoRA tuning technique, paired with well-suited rank settings, in yielding optimal results across a majority of tasks. 
PEFT-enhanced \sys attains a noteworthy 6\% accuracy boost over \texttt{TS-models}, while engaging a mere 0.0253\% of parameters for fine-tuning on average. 

Diving deeper, Figure~\ref{fig:peft_time} provides a comprehensive analysis of the impact of diverse PEFT techniques and associated hyper-parameters on the performance of PEFT-enhanced \sys.
In Figure~\ref{fig:peft_time}(a), the discernible trend showcases LoRA tuning as a standout performer, surpassing Prompt and Prefix tuning by 19\% and 52\% in terms of accuracy. 
Additionally, the fine-tuning process using LoRA mandates a mere 16.8 million trainable parameters, resulting in an exceptionally frugal training cost.
Figure~\ref{fig:peft_time}(b) offers further insights, indicating that selecting an appropriate LoRA rank value plays a pivotal role in propelling \sys towards heightened model accuracy while simultaneously minimizing trainable parameter size. For instance, with the LoRA rank set at 32, \sys attains a commendable accuracy of 98\% on the task, leveraging a mere 16.8 million trainable parameters.

\subsection{Runtime Cost}
\label{subsec:cost}






This subsection evaluates the storage, peak memory, latency, and energy consumption of running \sys and 50 \texttt{TS-models} on Jetson ORIN and Pixel 7 Pro.


\begin{figure}[t]
	\begin{subfigure}{.235\textwidth}
		\centering
		\includegraphics[width=\textwidth]{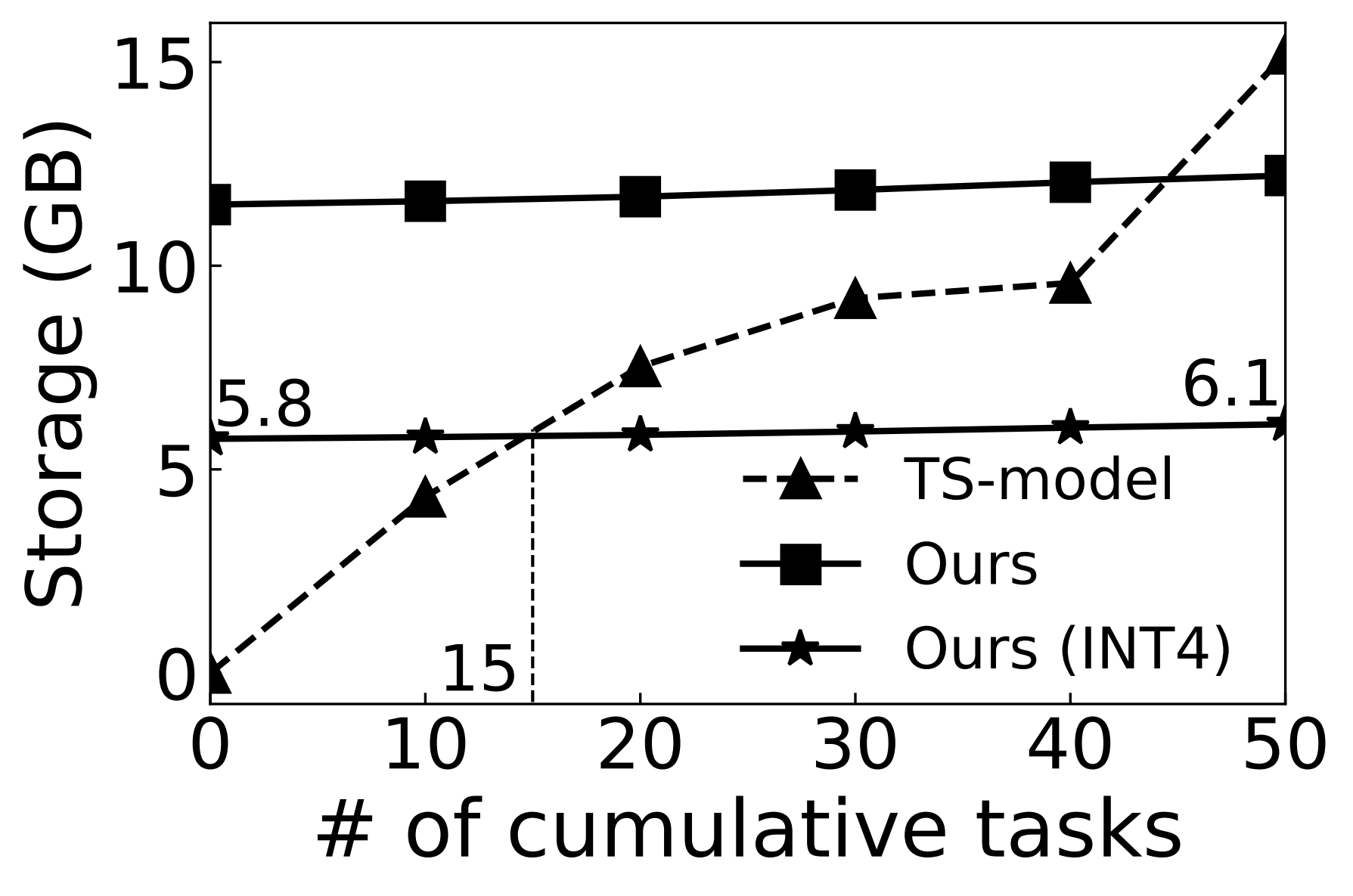}
		\caption{Storage}
	\end{subfigure}
	\begin{subfigure}{.235\textwidth}
		\centering
		\includegraphics[width=\textwidth]{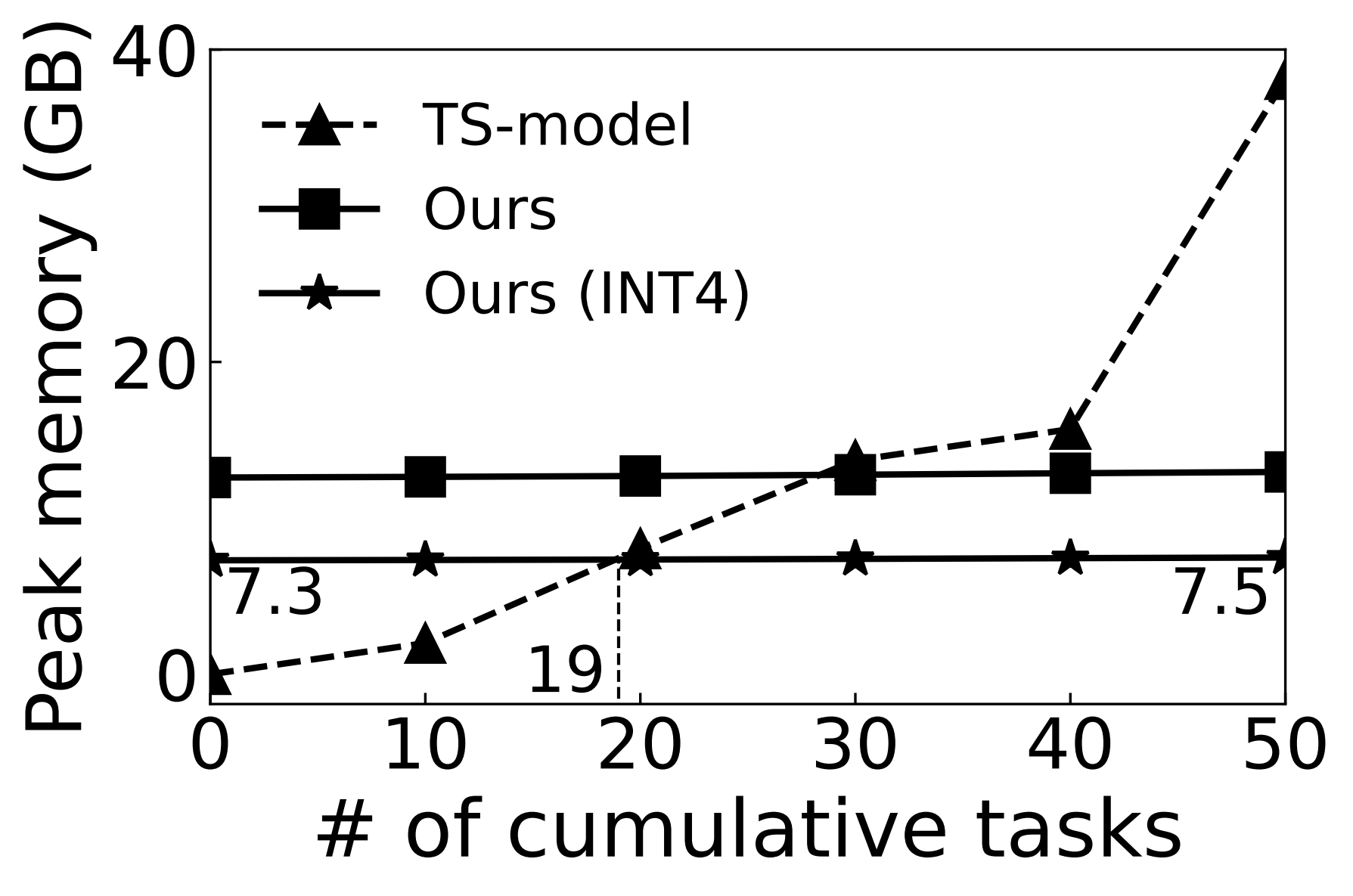}
		\caption{Peak memory}
	\end{subfigure}
	\caption{\sys's scalability analysis of storage and peak memory measured on Jetson ORIN.}
	\label{fig:cost-storage-footprint}
	\vspace{-12pt}
\end{figure}

\noindent \textbf{\sys is more storage-efficient when the model number scales out.}
Figure~\ref{fig:cost-storage-footprint}(a) presents a comparative analysis of storage between \sys and \texttt{TS-models} as the task count increases.
As observed, \sys's storage footprint is notably greater when serving a limited number of tasks compared to \texttt{TS-models}. 
However, the narrative changes as task diversity proliferates. With the deployment of a modest number of tasks (e.g. 15 tasks), the storage of \sys, specifically those equipped with INT4 quantization, outpaces that of \texttt{TS-models}. This trend intensifies as the number of tasks expands. 
Ultimately, \texttt{TS-models} surpass the storage allocation of INT4-based \sys, culminating at 15.2GB, signifying a substantial 2.5-fold escalation.
This underscores \sys's compelling storage scalability. 

\noindent \textbf{\sys is memory hungry, but is capable of holding more tasks for warm in-memory inference when task number scales out.}
Figure~\ref{fig:cost-storage-footprint}(b) shows that even when serving 50 tasks simultaneously, the cumulative peak memory usage of INT4-based \sys remains at a modest 7.5GB.
This constitutes a mere 2.7\% increase, while concurrently yielding a notable 5.1-fold reduction in peak memory consumption compared to \texttt{TS-models}.
This exceptional memory efficiency can be attributed to \sys's foundation design, which initially houses all requisite model parameters. Subsequently, the integration of new tasks necessitates only a marginal addition of fine-tuning parameters, typically amounting to less than 10MB each. The 7.5GB of peak memory cannot fit some mobile devices, but it is entirely affordable for many high-end smartphones with 12/16/32GB of RAM, like the Pixel 7 Pro we used. 
This underscores \sys's practicality and potential to be effectively deployed across a spectrum of devices.

\begin{figure}[t]
	\begin{subfigure}{.235\textwidth}
		\centering
		\includegraphics[width=\textwidth]{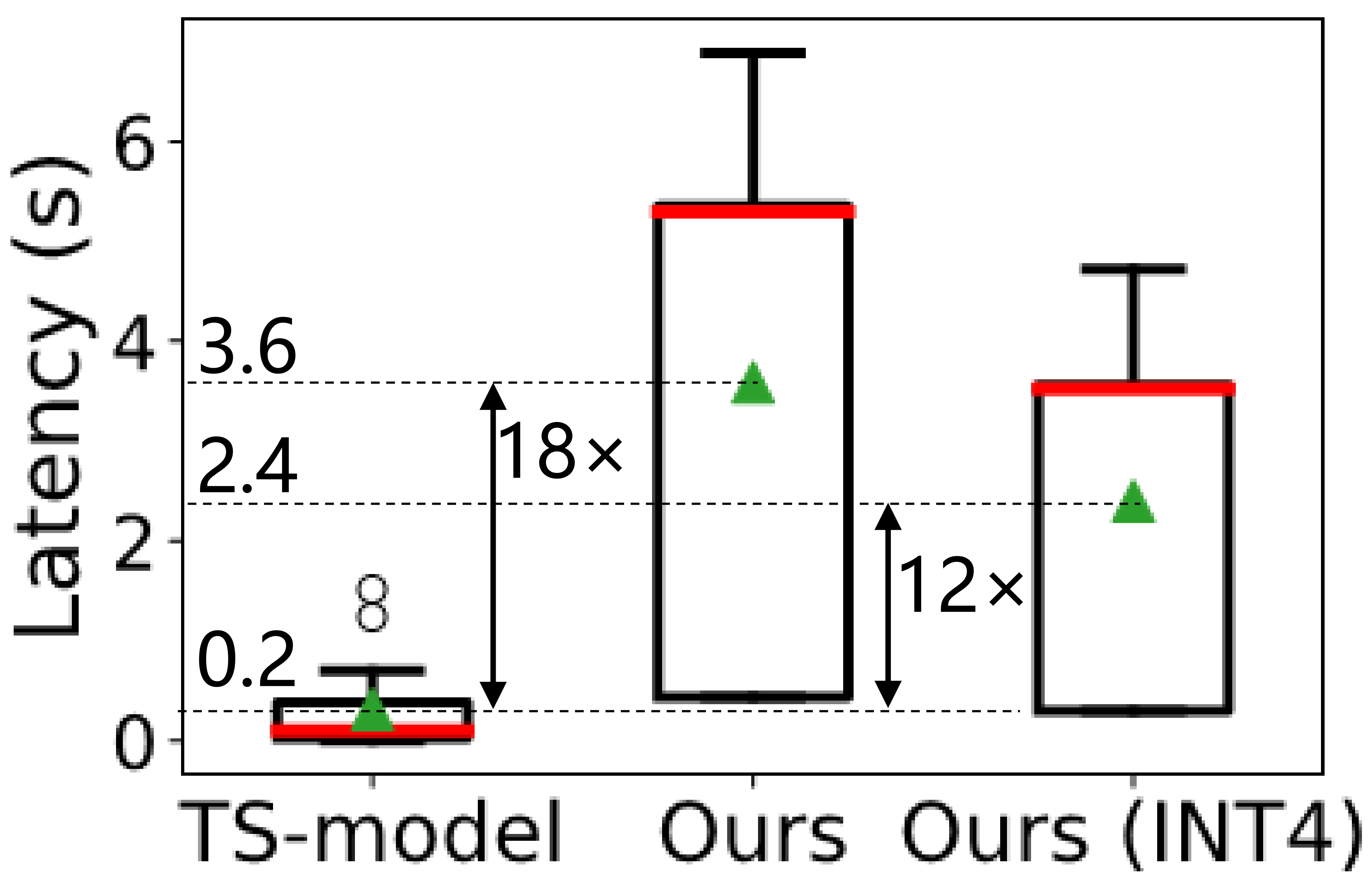}
		\caption{Average latency}
	\end{subfigure}
	\begin{subfigure}{.235\textwidth}
		\centering
		\includegraphics[width=\textwidth]{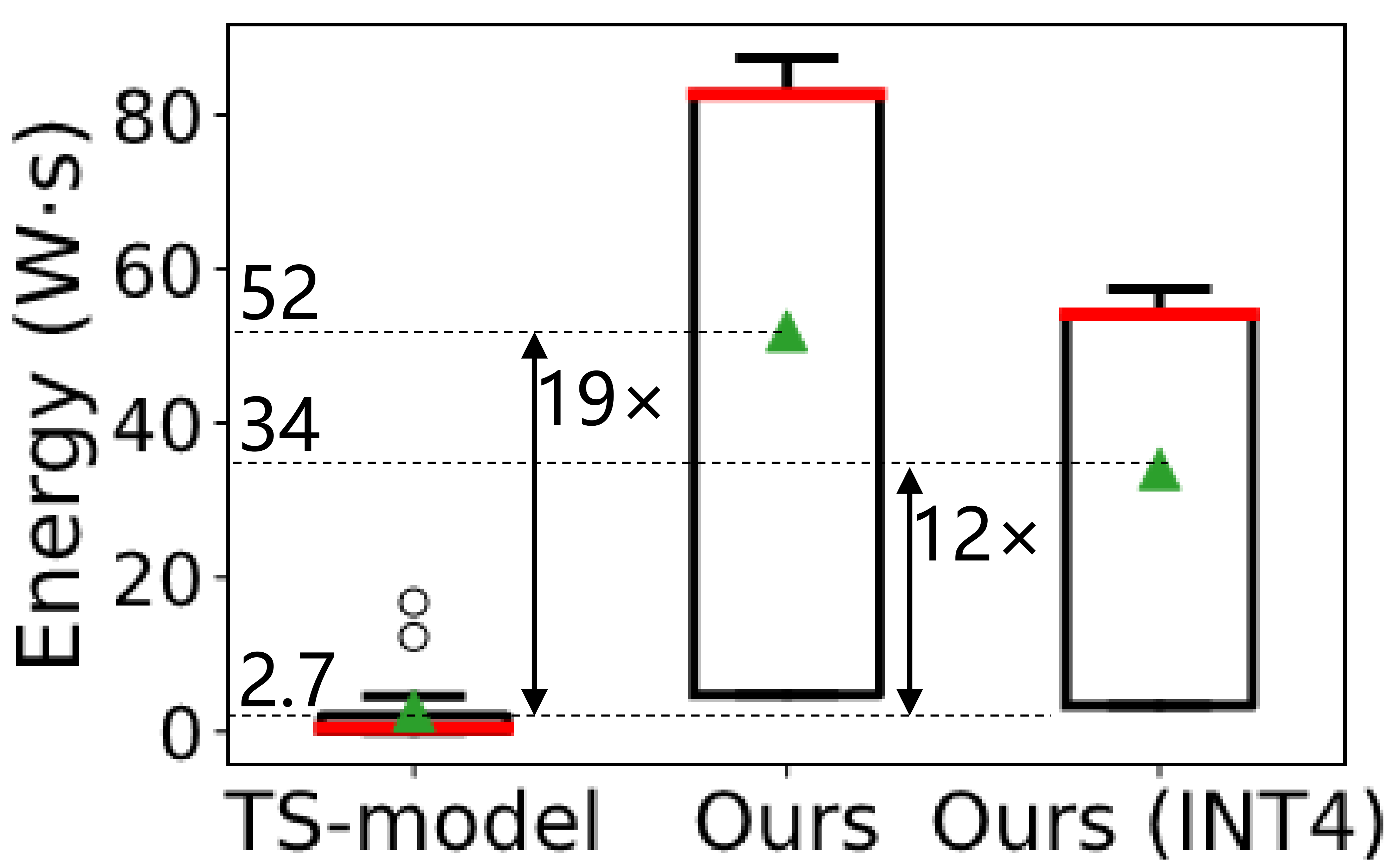}
		\caption{Average energy}
	\end{subfigure}
	\caption{\sys's runtime cost of latency and energy measured on Jetson ORIN (GPU).}
	\label{fig:cost-latency-energy}
	\vspace{-15pt}
\end{figure}
\noindent \textbf{\sys is 18$\times$ slower and incurs 19$\times$ more energy than \texttt{TS-models} on the same processor.}
Figure~\ref{fig:cost-latency-energy} provides a comparison of the inference latency and energy consumption between \sys and \texttt{TS-models} across the spectrum of 50 tasks. 
As observed, \sys using INT8-format LLaMA exhibits 12$\times$ and 19$\times$ (on average) higher inference latency and more energy consumption, compared to \texttt{TS-models}. 
While the transition to INT4 quantization offers a marginal amelioration, the performance gap remains significant—a respective 8$\times$ increase in latency and 12$\times$ surge in energy consumption compared to \texttt{TS-models}.
This substantial performance degradation is primarily due to \sys's substantial parameter count and intricate computational demands.


\begin{figure}[t]
	\begin{subfigure}{.235\textwidth}
		\centering
		\includegraphics[width=\textwidth]{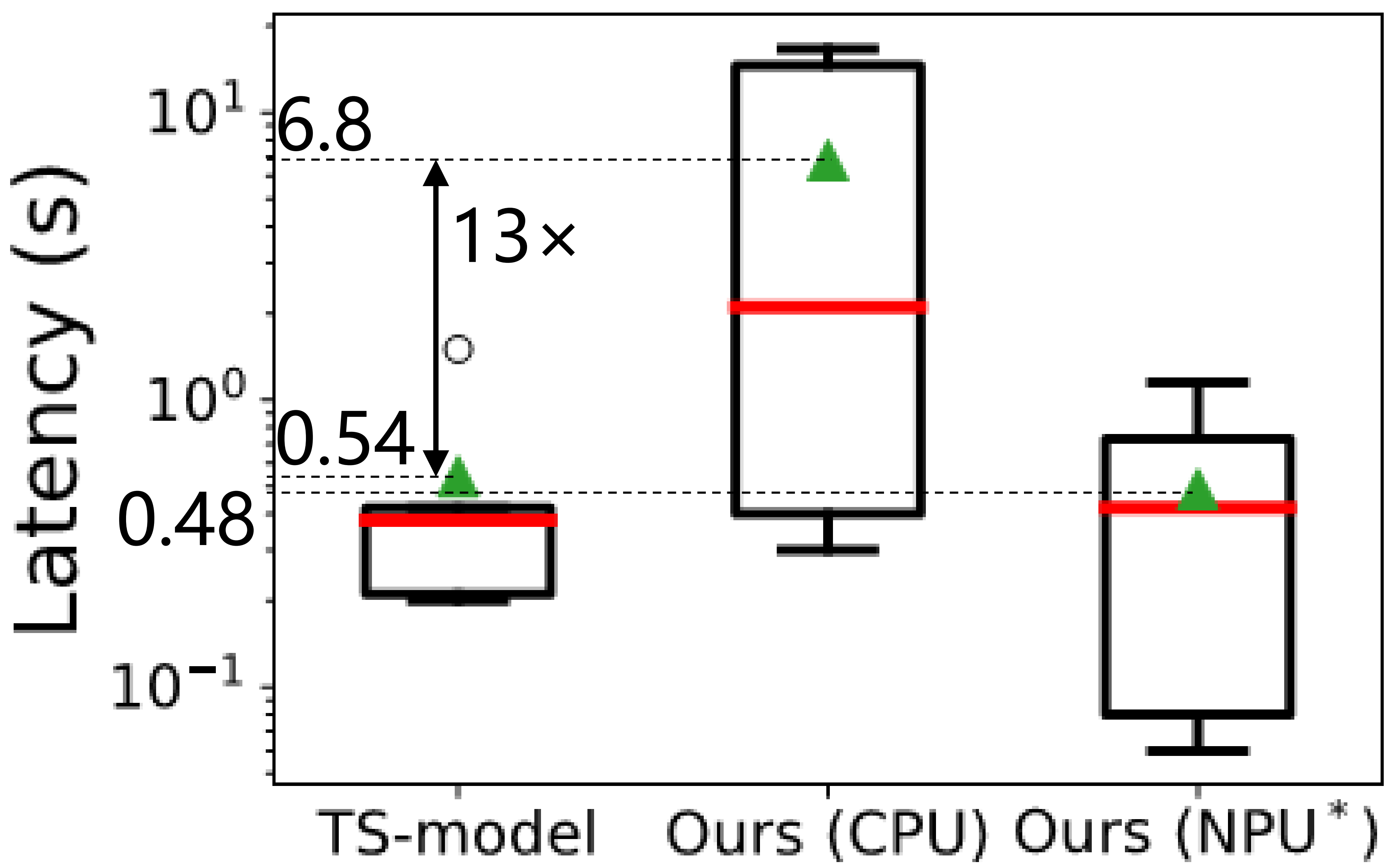}
		\caption{Latency}
	\end{subfigure}
	\begin{subfigure}{.235\textwidth}
		\centering
		\includegraphics[width=\textwidth]{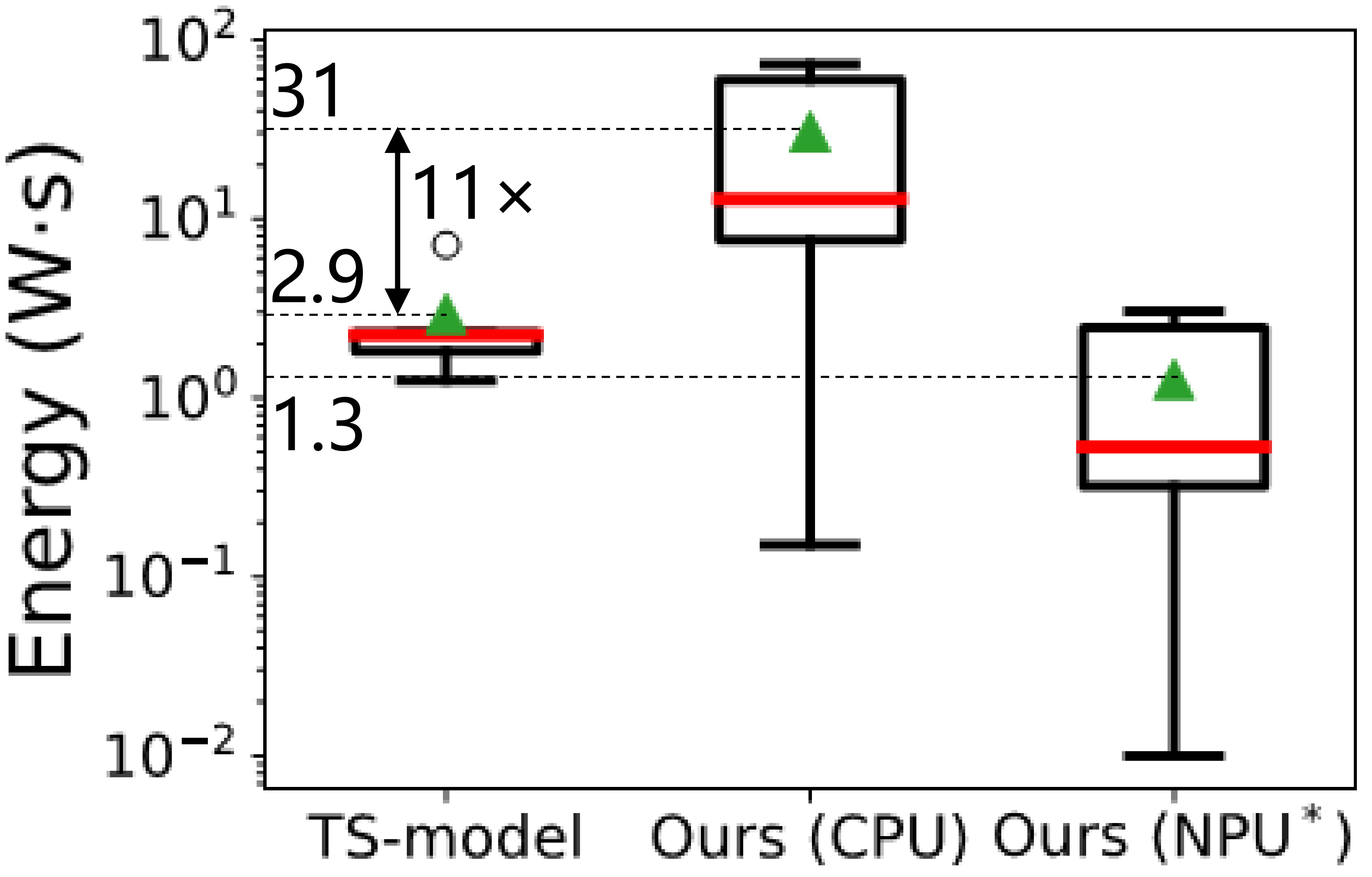}
		\caption{Energy}
	\end{subfigure}
	\caption{What-if cost analysis of latency and energy when running \sys on Pixel 7 Pro. \texttt{TS-model}: on CPU.}
	\label{fig:what-if}
	\vspace{-10pt}
\end{figure}
\noindent \textbf{\sys could get on par execution speed as \texttt{TS-models} if it can be deployed to run on the NPU.}
Figure~\ref{fig:what-if} provides a runtime cost comparison between \sys and TS-models on the CPU and NPU. 
We obtain the latency and energy of 50 tasks on CPU, denoted as TS-model and Ours (CPU).
As observed, \sys's inference latency and energy consumption on CPU are notably 13$\times$ and 11$\times$ (on average) higher than the TS-models. 

Given the substantial performance advantage of NPU over CPU as shown in $\S$\ref{subsec:bkgnd-heterNN}, we aim to evaluate the optimized runtime cost when deploying \sys on the NPU.
However, the NPU currently supports a limited set of operators (details in $\S$\ref{subsec:bkgnd-heterNN}) and cannot directly execute all components of \sys.
Similarly, the majority of TS-models cannot run directly on the NPU.
Therefore, we conduct a what-if analysis to estimate its runtime latency and energy consumption on NPU, denoted as Ours ($NPU^{*}$). 
This projection is achieved by leveraging the observed performance ratio from TS-models between the NPU and CPU, as discussed in Section $\S$\ref{subsec:bkgnd-heterNN}. 
Subsequently, utilizing the measured performance of \sys on the CPU, we can derive its estimated performance on NPU.

As observed, NPU-enabled \sys achieves an average latency of 0.48s and energy consumption of 1.3J, which are even 11.1\% and 55.2\% lower than TS-models on CPU. 
Furthermore, we delve into the architectural intricacies of \sys to analyze the latency breakdown performance in Table~\ref{tab:whatif-npu}. 
From this table, we observe that the latency optimization bottleneck for \sys lies in the time taken by the IMG-encoder and the generation of the first token by the backbone, which is approximately 2.1s and 6.3s, respectively. 
These components collectively account for 31\% and 93\% of \sys's average latency (6.8s in Figure~\ref{fig:cost-latency-energy}(a)).
However, the other components could exhibit near real-time inference (less than 100ms) if being deployed on NPU.
These achievements demonstrate that, if \sys can be accelerated on NPU, it could get on par execution speed and energy consumption with TS-models.
The aim of comparing \sys's NPU performance with TS-models on CPU isn't to claim superiority, but to showcase how future NPU support can enhance efficiency for mobile foundation models.

\begin{table}[]
    \centering
    \resizebox{\columnwidth}{!}{%
    \begin{tabular}{|c|c|cc|}
    \hline
    \multirow{2}{*}{\textbf{Tasks}}                                                       & \multirow{2}{*}{\textbf{Path}} & \multicolumn{2}{c|}{\textbf{Latency (s)}}                       \\ \cline{3-4} 
                                                                                            &                                & \multicolumn{1}{c|}{\textbf{CPU}}               & \textbf{NPU*} \\ \hline
    Image classification                                                                  & Path-3                         & \multicolumn{1}{c|}{IMG\_enc: 2.10}             & 0.11          \\ \hline
    Audio classification                                                                  & Path-3                         & \multicolumn{1}{c|}{AUD-I\_enc: 0.28}           & 0.014         \\ \hline
    \multirow{2}{*}{Question answering}                                                   & \multirow{2}{*}{Path-2}        & \multicolumn{1}{c|}{First token: 6.34}          & 0.32          \\ \cline{3-4} 
                                                                                            &                                & \multicolumn{1}{c|}{Sequent tokens: 0.24/token} & 0.012/token   \\ \hline
    \multirow{2}{*}{\begin{tabular}[c]{@{}c@{}}Visual \\ question answering\end{tabular}} & \multirow{2}{*}{Path-1}        & \multicolumn{1}{c|}{First token: 6.47}          & 0.32          \\ \cline{3-4} 
                                                                                            &                                & \multicolumn{1}{c|}{Sequent tokens: 0.25/token} & 0.013/token   \\ \hline
    Text-to-speech                                                                        & Path-4                         & \multicolumn{1}{c|}{TTS\_dec: 0.82}             & 0.041         \\ \hline
    \end{tabular}
    }
    \caption{An in-depth what-if cost analysis of latency when running \sys on Pixel 7 Pro. NPU*: \sys's estimated latency based on the NPU acceleration rate of \texttt{TS-model} if it can be deployed on NPU.}
    \label{tab:whatif-npu}
    \vspace{-15pt}
\end{table}

\subsection{Model Architecture Simplicity}
\label{subsec:simplicity}


\begin{figure}[t]
	\begin{subfigure}{.235\textwidth}
		\centering
		\includegraphics[width=\textwidth]{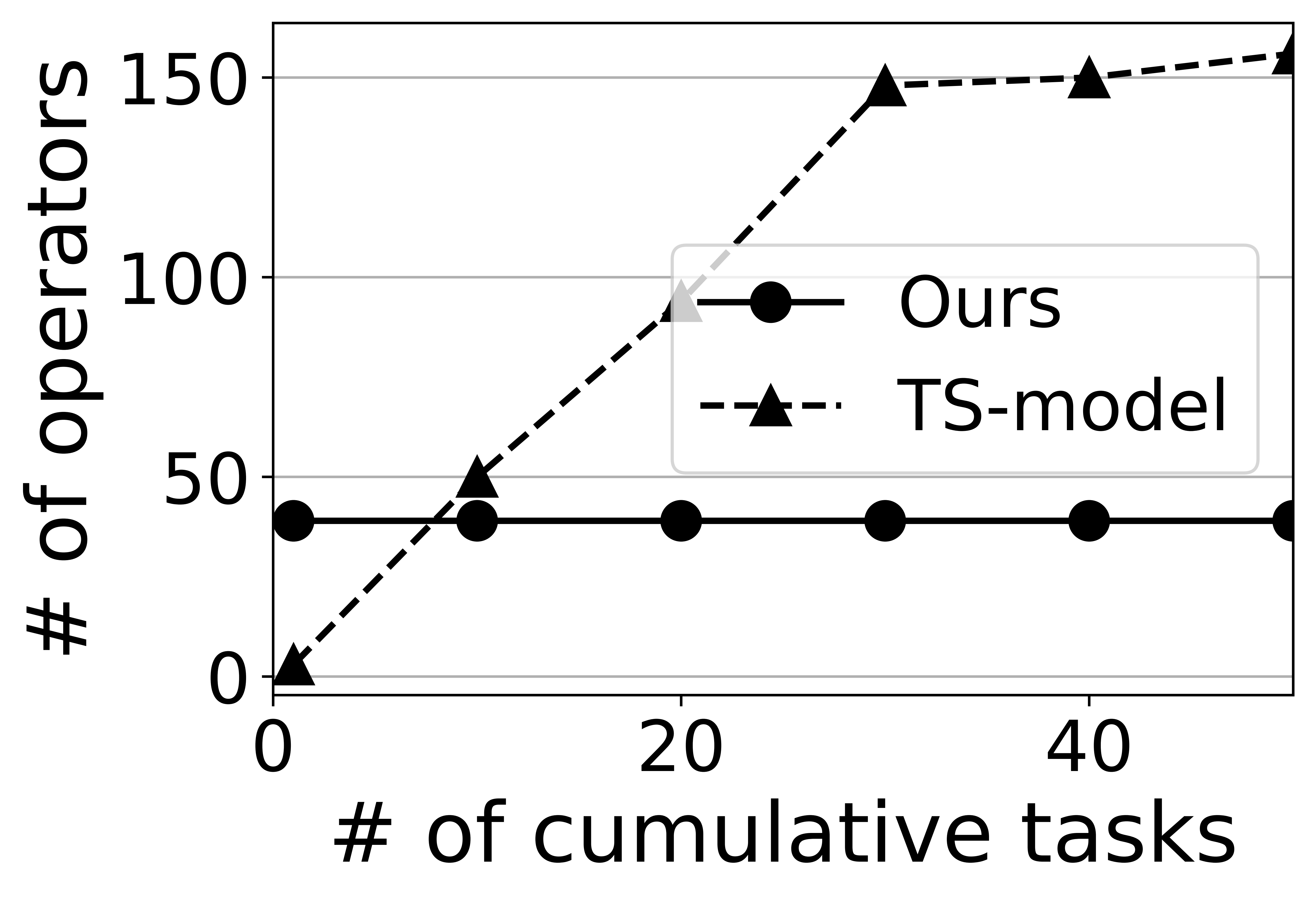}
		\caption{Total operators}
	\end{subfigure}
	\begin{subfigure}{.235\textwidth}
		\centering
		\includegraphics[width=\textwidth]{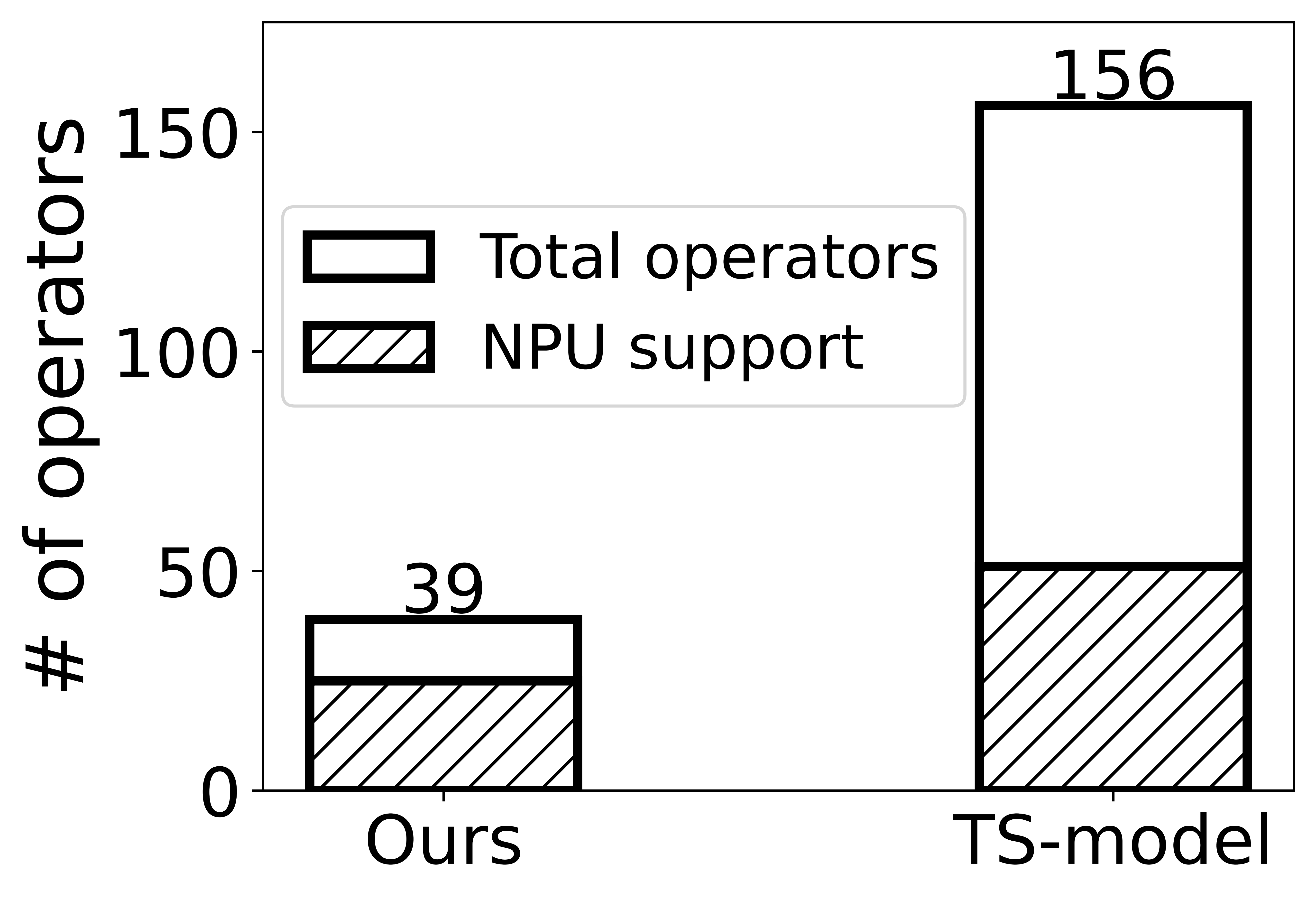}
		\caption{NPU supported operators}
	\end{subfigure}
	\caption{Simplicity analysis of \sys's operators.
	}
	\label{fig:simplicity-operations}
	\vspace{-15pt}
\end{figure}



\noindent \textbf{\sys's architectural design is much simpler and cleaner in terms of NN operators, therefore could greatly simplify accelerator design.}
Figure~\ref{fig:simplicity-operations}(a) shows that the number of operators in the \texttt{TS-models} increases rapidly with the growth in the number of tasks. 
Notably, as the task spectrum broadens to encompass 50 tasks, the number of operator types culminates at 156. 
In contrast, \sys engages a mere 39 operator types, encompassing both foundation model and task-specific "adapters".
Furthermore, Figure~\ref{fig:simplicity-operations}(b) undertakes a granular exploration of NPU supported operators for both \sys and \texttt{TS-models}.
It underscores that only 51 out of 156 operators in \texttt{TS-models} are supported by the NPU, with more than 2/3 of the operators unable to fully run on the NPU; 
But for \sys, NPU-supported operators account for 64\%.
This phenomenon emanates from \sys's transformer-based architecture, which inherently involves quantity and NPU friendly operators, thus enhancing operational efficiency~\cite{pati2022demystifying}.

\subsection{Novel Application with \sys}
\label{subsec:usecase}

\begin{figure}[t]
	\centering
	\includegraphics[width=0.47\textwidth]{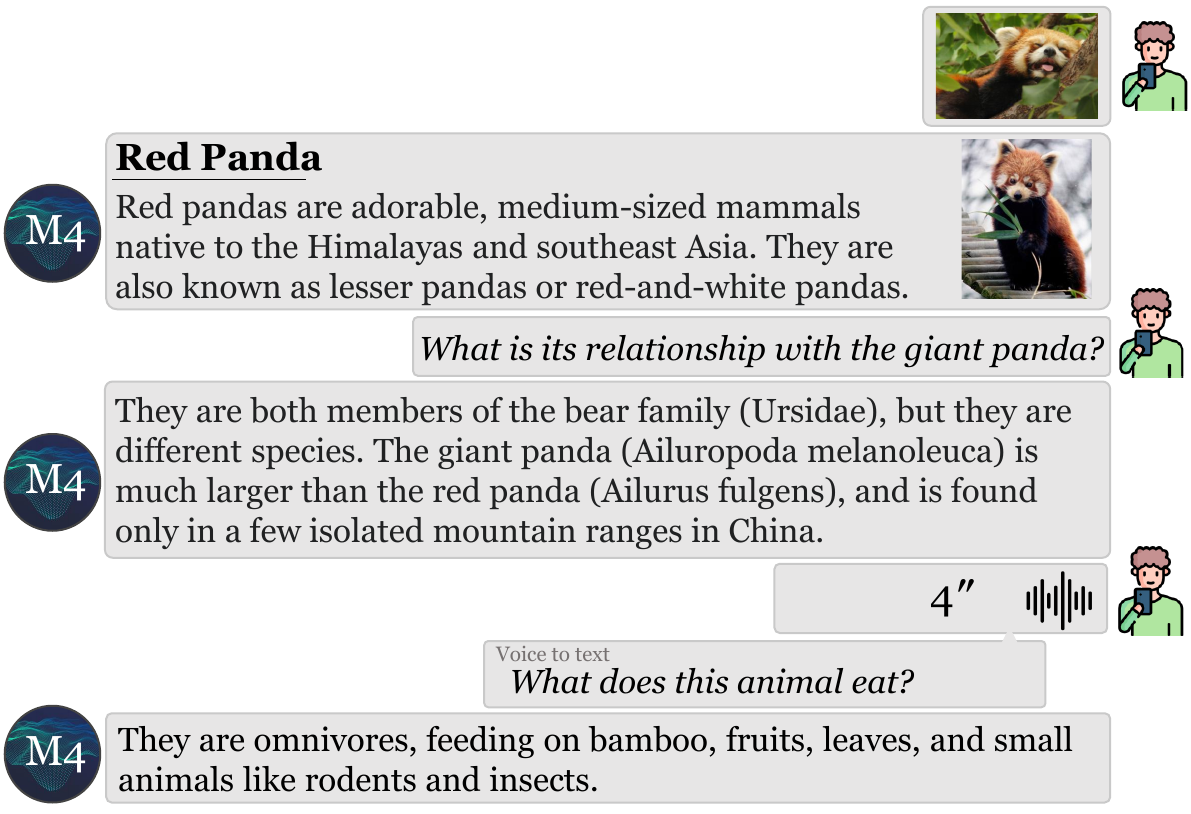}
	\vspace{-5pt}
	\caption{A demo of \sys: multimodal chat.}
	\label{fig:usecase}
	\vspace{-5pt}
\end{figure}

\textbf{\sys enables complex, unpresent mobile applications.}
Based on our proposed \sys, we build a demo of a multimodal chat use case as shown in Figure~\ref{fig:usecase}.
Users engage in multi-turn chats with the M4 client using multimodal inputs such as images, text, and audio, thereby obtaining precise and tailored answers that meet their requirements.
This multimodal computing capability is also crucial for the recent popular mobile agents~\cite{li2024personal}.
We build this prototype system of \sys based on the architecture depicted in Figure~\ref{fig:MobileFM-archi}.
It first aligns the contents of image, text, and audio by converting multimodal input data into a unified representation.
Then, it encapsulates abundant knowledge to understand complex embedded data, perform task-specific inference, and generate the required information.
This innate capability for multimodal processing harbors the potential to significantly enrich the landscape of mobile applications. 
    \section{Related Work}\label{sec:related}

\noindent \textbf{Foundation models.} Building one-for-all foundation models to serve generic AI tasks has been a primary research goal of the machine learning community.
The recent advancements of LLMs \cite{touvron2023llama, touvron2023llama2, GPT4, zeng2022glm}, multimodalities alignment \cite{girdhar2023imagebind, tang2023Codi, su2023pandagpt, najdenkoska2023meta}, and parameter-efficient training methods \cite{peft2023, zaken2021bitfit, yang2023caas} have shed lights on this challenging goal.
For instance,  ImageBind \cite{girdhar2023imagebind} and CoDi \cite{tang2023Codi} focus on how to align the embeddings of each modality, and PandaGPT \cite{su2023pandagpt} further attempts to compose semantics from different modalities naturally based on LLaMA \cite{touvron2023llama}.
However, there have been no efforts like \sys that try to fit extremely diversified AI tasks into one model.
Meanwhile, \sys leverages the most state-of-the-art pre-trained LLMs to reuse the wisdoms as well as the investments from the ML community\&industry.
The concurrent work NExT-GPT~\cite{wu2023next} shares a similar architecture as \sys.
Nonetheless, \sys introduces two distinctive contributions:
(1) It marks the inaugural proposal of a transformer-based N-1-M architecture, aiming to curtail resource costs in any-to-any modal generation;
(2) Its innovative multi-path execution design is tailored to enhance compatibility with highly diversified mobile AI tasks.

\noindent \textbf{Hardware-system-algorithm co-design for mobile AI.}
AI workloads are highly compute-intensive and exhibit analogous patterns, therefore is better to be accelerated domain-specific accelerator (e.g., NPUs).
For instance, SpAtten \cite{wang2021spatten} and Sanger \cite{lu2021sanger} focus on how efficient algorithm-architecture co-designs can reduce sparse attention computation and memory access. 
Besides, QNAS \cite{lin2019neural} and NAAS \cite{lin2021naas} focus on composing highly matched neural-hardware architectures by jointly searching for neural network architecture, accelerator architecture, and compiler mapping.
However, all prior literature makes tradeoffs between the ubiquity of operator support and the performance, instead of for a foundation model that can serve generic AI tasks itself.
The vision of the mobile foundation model could open a new research domain for cross-layer co-design of mobile AI.
There have been  preliminary attempts~\cite{xu2024survey,yi2023edgemoe,xu2023federated} to alleviate the huge resource cost of large foundation models for devices. Those work are orthogonal to this work.

\noindent \textbf{Managing AI as a mobile system service.}
AI has been a ubiquitous workload on mobile devices, and managing it at a system aspect (instead of individual app) could facilitate OS-wise runtime scheduling and software deployment.
Some early studies \cite{xu2018deepcache, zhao2022tsn, wang2019understanding, eslami2021robustness, zhang2022comprehensive} attempt to mitigate the severe fragment across different libs in the mobile DL ecosystem.
Google introduced a unified ML interface NNAPI \cite{NNAPI2017} into Android in 2017, to relieve the gap between heterogeneous mobile processors and user-defined ML frameworks.
Compared to the above work, \sys takes another giant step further that mobile devices shall manage a foundation model for each ML task and expose it as firmware.

    \section{Limitations and Future Work}
This study has several potential limitations.
(1) \benchmark's results are evaluated on datacenter GPU (NVIDIA A100) and edge GPU (Jetson Orin), lacking assessment on mobile devices.
It's mainly due to the highly diverse code implementation of baseline models and the huge time span of evaluating \sys on large test dataset.
There might exist performance gap between different hardware architectures.
Yet, the comparison is fair as both baseline models and \sys are evaluated on the same hardware.
In fact, due to the simpler and cleaner architecture of \sys, it would be much easier to design accelerator to support \sys with high precision.
(2) \sys underperforms baseline models on certain ML tasks. This unveils the limitation of existing pre-trained foundation models, e.g., translation.
On the one hand, we do not expect \sys to be able to solve all mobile AI tasks in the near future; it could co-exist with traditional DNNs that run on mobile CPU.
On the other hand, the LLM capacity is still fast evolving: from LLaMA-1/2 used in this study, to the Mistral-7B~\cite{jiang2023mistral} that ranks higher even than LLaMA-13B.
Such continuous improvement endeavors our vision with much confidence.

To be noted, \sys is the very first step towards the vision of mobile foundation model.
We believe it could potentially revolutionize the mobile AI landscape and open a new research domain.
However, to fully realize the vision, there are a few key designs to be explored.
For instance:
(1) \textit{Foundation model design}: As a preliminary prototype, \sys is currently built atop off-the-shelf, separately pre-trained LLMs from Internet instead of being tailored for mobile devices.
Therefore, it is still highly inefficient in terms of accuracy and model parameter size.
With enough resources (GPUs and data), hardware vendors can build a more compact mobile foundation model that is expected to deliver significantly higher accuracy with lower runtime cost than \sys.
(2) \textit{Accelerator design}: fine-tuning for downstream tasks generates small ``adapters'' that are inserted into the mobile foundation model. The NPU better has the flexibility to run those adapters as well; otherwise the inference must involve CPU/GPU computation and data movement overhead.
Fortunately, the adapters have simple structure (e.g., linear matrix operations) and very few weights.
(3) \textit{FM upgradating}: the foundation model capacity could evolve with better architecture/weights as shown in $\S$\ref{subsec:accuracy}.
Yet the adapters trained for the old foundation model cannot work with the new one.
We therefore need a unified interface between LLMs and adapters to allow them to evolve independently without interfering with each other.

\section{Conclusions}\label{sec:conclusions}
We envision a mobile hardware-OS co-managed multimodal foundation model that is exposed to mobile applications as a system service to serve almost every on-device AI task.
We design and prototype the first such model using off-the-shelf LLMs.
Evaluated on a comprehensive benchmark consisting of 50 representative mobile AI tasks, \sys shows good accuracy, better scalability and reduced runtime cost.

    \section{Acknowledges}
The authors thank the anonymous reviewers and the shepherd for their insightful feedbacks. This work was supported by NSFC (61921003, 62102045).
    
    \balance
    \bibliographystyle{plain}
    \bibliography{ref-mwx,ref-cdq,ref-yc,ref-yuan,ref-bench}

\end{document}